\documentclass[journal]{IEEEtran}
\usepackage[utf8]{inputenc}
\usepackage{multicol}
\usepackage{amsmath}
\usepackage{multirow}
\usepackage{amssymb}
\usepackage{verbatim}
\usepackage{booktabs}
\usepackage{graphicx}
\usepackage{longtable}
\usepackage{blindtext}
\usepackage{float}
\usepackage{array}
\usepackage{xcolor, colortbl}
\definecolor{Gray}{gray}{0.9}
\usepackage{etoolbox}
\usepackage[normalem]{ulem} 
\makeatletter
\patchcmd{\@makecaption}
  {\scshape}
  {}
  {}
  {}
\makeatother

\usepackage{cite}

\usepackage{flushend}

\usepackage[switch,pagewise]{lineno}

\usepackage{enumitem}
\setlist[itemize]{noitemsep, topsep=1pt, leftmargin=*}

\author{\IEEEauthorblockN{Salma Abdel Magid,
Francesco Petrini, and Behnam Dezfouli\IEEEauthorrefmark{1}\thanks{\IEEEauthorrefmark{1}Corresponding Author}}\\
\IEEEauthorblockA{Internet of Things Research Lab, Department of Computer Science and Engineering, Santa Clara University, USA
\\
  \texttt{\small sabdelmagid@scu.edu, fpetrini@scu.edu, bdezfouli@scu.edu}
}}

\title{Image Classification on IoT Edge Devices:\\ Profiling and Modeling}

\date{August 2018}

\begin{document}


\maketitle
\begin{abstract}
With the rise of powerful, low-cost IoT systems, processing data closer to where the data originates, known as edge computing, has become an increasingly viable option.
In addition to lowering the cost of networking infrastructures, edge computing reduces edge-cloud delay, which is essential for mission-critical applications.
In this paper, we show the feasibility and study the performance of image classification using IoT devices.
Specifically, we explore the relationships between various factors of image classification algorithms that may affect energy consumption, such as dataset size, image resolution, algorithm type, algorithm phase, and device hardware. 
In order to provide a means of predicting the energy consumption of an edge device performing image classification, we investigate the usage of three machine learning algorithms using the data generated from our experiments. 
The performance as well as the trade-offs for using linear regression, Gaussian process, and random forests are discussed and validated.
Our results indicate that the random forest model outperforms the two former algorithms, with an R-squared value of 0.95 and 0.79 for two different validation datasets. 
The random forest also served as a feature extraction mechanism which enabled us to identify which predictor variables influenced our model the most.



\end{abstract}

\section{Introduction}

Researchers at Gartner estimate that there will be 20 billion IoT devices connected to the Internet by 2020 \cite{Gartner}. 
The burgeoning of such devices has sparked many efforts into researching the optimal device design. 
Since most IoT devices are constrained in terms of processing power and energy resources, the traditional approach has been to transmit data generated by the device to a cloud platform for server-based processing. 
Although cloud computing has been successfully employed, it is sometimes not desirable due to concerns about latency, connectivity, energy, privacy, and security \cite{MICROSOFT,FOGP1,FOGP2}.

To overcome these concerns, edge and fog computing have emerged.
These architectures aim to push processing capabilities closer to the IoT devices themselves, which is specifically possible given their significant increase in processing power.
For example, the archetype of modern IoT devices, the Raspberry Pi 3, offers a quad-core processor with 1GB of RAM for only \$30.
The reduction in latency offered by utilizing such devices in edge and fog computing is critical to the success of applications such as object detection and image classification.
These applications are used in mission-critical systems such as autonomous vehicles, surgical devices, security cameras, obstacle detection for the visually-impaired, rescue drones, and authentication systems \cite{IotApplications,IoTWearables,IoTDevices}. 
However, these tasks consume a considerable amount of energy. 
Thus, it is especially important to understand the relationship between these algorithms and their respective energy consumption to efficiently utilize the IoT device's power resources. 
This is particularly important due to two reasons: First, many of these IoT devices work in a duty-cycled fashion. 
They are triggered when an external event happens, perform a processing task, and transition to sleep mode once the task completes.
A sample scenario is a security camera that captures an image when motion is detected.
Another example could be a flood monitoring system that captures images of a river when the water level is beyond a certain threshold to detect the type of debris being carried by the water.
Enhancing energy efficiency is essential for these types of applications, especially when they are battery powered or rely on energy harvesting technologies \cite{amirtharaj2018profiling}.
The second important motivation towards energy profiling and enhancement is to reduce carbon emissions. 
According to a study published by the Centre for Energy Efficient Telecommunications, the cloud was estimated to consume up to 43 TWh in 2015, compared to only 9.2 TWh in 2012, an increase of 460\% \cite{CEET}. 
This is roughly equivalent to adding 4.9 million cars to the roads.
Given the dramatic impact of inefficient energy management, it has become important to ensure that the most intensive of tasks, especially image classification, are using the appropriate resources and minimizing their energy consumption footprint.

Various machine learning (ML) algorithms, offering different accuracy and complexity, have been proposed to tackle the challenges of image classification.
Despite their exceptional accuracy, they require high processing power and large storage.
For example, some state-of-the-art neural network architectures, such as AlexNet \cite{AlexNet}, GoogLeNet\cite{GoogleNet}, and ResNet \cite{ResNet} require over a million parameters to represent them and more than a billion multiply and accumulate computations (MAC) \cite{IoTEnergy}. 
Each MAC operation is generally associated with a number of memory accesses. 
In the worst case scenario, where there is no data re-use, each operation requires 3 reads and 1 write to memory. 
The simplest neural network from the aforementioned models requires around 2172M memory reads and 724M memory writes. 
Since these operations consume a considerable amount of processing power, the energy consumption of these algorithms might not meet the requirements of various application scenarios.
However, the overall energy consumption can be reduced if the number of operations performed by these algorithms is also reduced.
This is possible through various approaches such as reducing image resolution, reducing dataset size, and choosing the algorithm that addresses the application requirements without introducing additional processing overhead. 
For example, ResNet-50 processing a 224$\times$224$\times$3 image uses around 7 billion operations per inference \cite{ARM}. 
Running this neural network on a 160$\times$160 image would almost halve the number of operations and double the speed, immensely reducing the energy consumption.
In terms of algorithm selection, some algorithms are better suited for servers (where there is a wider variety of accessible resources), whereas others can perform well on IoT devices. 
If, for example, the energy consumed to classify a single image on the device was considerably less than the energy consumed to transmit the image to the cloud and receive the result, then, as one scales, it becomes advantageous to compute locally.

There have been research efforts to deliver preliminary observations as to how resource-constrained embedded devices perform while executing ML algorithms. 
Cui et al. \cite{Wanlin} used a Raspberry Pi 2 as a gateway and a commercial Intel SR1560SF server. 
They found a strong relationship between energy and data size.
In addition, they found that for some scenarios, the gateway, which employs a low-power processor, performs data processing tasks using a lower amount of energy compared to the server over a long period of time.
However, their study focused on how ML algorithms perform for general tasks and generated a model to predict energy consumption solely based on data size. 
Unfortunately, they did not consider how the type and phase of the algorithm or how specific data characteristics, such as image resolution, impact performance. 
Carbajales et al. \cite{Carbajales} investigated the power requirement of IoT monitoring and sensing on a Raspberry Pi 2 for a smart home application. 
Their goal was to present a user-friendly visualization of energy consumption across several single board computers (SBCs) including the Raspberry Pi 2B and the BeagleBone Black.
Their data processing was limited to time-scaling, averaging, summing, and rounding with no consideration for more complex processing such as ML. 
In addition, they did not propose any method to predict or forecast the energy requirements of the system. 
Lane et al. \cite{Lane} characterized neural network algorithms for various embedded devices including wearables and smartphones. 
They chose Nvidia Tegra, Qualcomm Snapdragon, and Intel Edison, and measured execution time and energy consumption for each.  
Out of the four deep learning architectures, two were used for object detection, namely AlexNet and Street View House Numbers (SVHN). 
While AlexNet has seen state-of-the-art accuracy and can distinguish more than 1,000 object classes, SVHN has a more narrow use case: extracting numbers from noisy scenes. Although this research incorporated deep learning, it did not include analysis of how the data characteristics (such as image resolution) influenced the energy consumption. To summarize, despite the insights provided by the aforementioned research efforts into performance, in terms of duration and energy, none of them have investigated the relationship between image input data versus energy, duration, and accuracy. 
Furthermore, these studies did not provide a useful technique for predicting the energy consumption when multiple parameters are taken into account.

The contributions of this paper are two-fold.
First, we identify and characterize how each individual factor of image classification can affect energy cost, duration, and accuracy. 
This will equip the research community with the tools necessary to make an informed decision about the design of their edge/fog systems in a way that balances cost with performance. 
Second, we present a reliable method for predicting the energy consumption of a system without needing to construct and measure data from a prototype. 
More specifically, in this paper:

\begin{itemize}

    \item [--] We analyze and visualize the relationships between energy consumption, duration, and accuracy versus dataset size, image resolution, algorithm type, algorithm phase (i.e., training and testing), and device type, when executing ML algorithms on IoT devices.
    The machine learning algorithms we used in this study are support vector machines (SVM), k-nearest neighbors (k-NN), and logistic regression. These algorithms were selected based on their popularity for image classification, as well as their abundant implementations across several frameworks.
    We chose the Raspberry Pi 3 (RPi) and the BeagleBone Black Wireless (BB) because they are widely used by the IoT community.
    We found that despite the BB's access to lower-voltage DDR3L RAM, which has twice the clock speed and transfer rate potential of the RPi's RAM, it generally always took significantly longer for the BB to perform experiments, ultimately leading it to consume more energy.
    This discrepancy, in part, is credited to the RPi's CPU which, despite being unable to utilize all four of its cores for some experiments, still has a 20\% faster clock speed than that of the BB.
    We present evidence that suggests increasing image resolution serves only to increase energy consumption while providing minimal benefit to accuracy.
    For example, using the RPi, we find that increasing the resolution of images by 40\% for datasets of size 300 and 1500 results in an average increase in processing time of 191\% and 217\%, and an average increase in energy of 208\% and 214\%, respectively.
    Despite these significant increases in energy consumption, the accuracy for the same datasets is \textit{decreased} by 3.64\% and 4.64\%, respectively, suggesting that, in general, for small datasets it is not beneficial to increase image resolution.
    Additionally, we conducted experiments utilizing the RPi's multi-core functionality and compared the results with the corresponding single-core data. In this way, we found that using multiple cores provided many benefits including a 70\% and 43\% reduction in processing time as well as a 63\% and 60\% decrease in energy consumption for k-NN and logistic regression, respectively.

    \item [--] Since energy measurement is a lengthy process and requires the use of an accurate power measurement tool, we utilize our experimental data to present a novel energy prediction model. 
    In our attempts to generate the most accurate model, we used three ML algorithms: multiple linear regression, Gaussian process, and random forest regression.
    After applying the ML algorithms to the validation datasets, random forest regression proved to be the most accurate method, with a R-squared value of 0.95 for the Caltech-256 dataset and 0.79 for the Flowers dataset.
    The proposed model facilitates decision making about the target hardware platform, ML algorithm, and adjustments of parameters, based on the application at hand.
\end{itemize}

Table \ref{key_terms} shows the key terms and notations used in this paper.
The remainder of the paper is organized as follows. 
Section \ref{MethodologySection} discusses the methodology of our experiments. 
Section \ref{ResultsAnalysisSection} presents experimentation results and provides a list of guidelines for the purpose of maximizing performance and system longevity.
Section \ref{ModelingRFSection} describes our proposed method to generate a random forest model capable of predicting energy consumption.
The paper concludes in Section \ref{ConclusionSection} by summarizing our findings and highlighting future work directions.

\begin{table}[]

\caption{Key Terms and Notations}
\begin{center}
\begin{tabular}{|c||l|}

\hline
Term & Description \\ \hline \hline
k-NN & \begin{tabular}[c]{@{}l@{}} k-Nearest Neighbors\end{tabular} \\ \hline
SVM & \begin{tabular}[c]{@{}l@{}} Support Vector Machine\end{tabular} \\ \hline
LOG & \begin{tabular}[c]{@{}l@{}} Logistic Regression\end{tabular} \\ \hline
Resolution & \begin{tabular}[c]{@{}l@{}}The square dimensions of an image measured\\ in pixels.\end{tabular} \\ \hline
Phase & \begin{tabular}[c]{@{}l@{}}The phase of the ML algorithm\\ (i.e., training or testing)\end{tabular} \\ \hline
\#Images & \begin{tabular}[c]{@{}l@{}}The number of images present in a dataset\\ (e.g., 300, 600, 900, 1200, 1500)\end{tabular} \\ \hline
\#Classes & \begin{tabular}[c]{@{}l@{}}The number of classes belonging to a specific\\ dataset (e.g., 2, 7, 10)\end{tabular} \\ \hline
RMSE & \begin{tabular}[c]{@{}l@{}} Root Mean Square Error: a measure of average\\ deviation between data points and the trend line\end{tabular} \\ \hline
$R^{2}$ & \begin{tabular}[c]{@{}l@{}} A measure of how closely data fits  to a regression\\ line\end{tabular} \\ \hline
\end{tabular}

\end{center}
\label{key_terms}

\end{table}

\section{Methodology}
\label{MethodologySection}

In this section, we present the main components of our measurement methodology, including hardware platforms, the power measurement tool, the ML algorithms, and the datasets.

\subsection{Hardware Platforms}

In order to maximize the relevance and applicability of our model, we selected hardware devices that are widely adopted by the IoT community.
Recent surveys suggest that the RPi is the most popular single board computer (SBC) \cite{2014,2016}.
The RPi contains a 1.2GHz quad-core ARM Cortex-A53 BCM2837 processor and 1 GB of DDR2 SDRAM \cite{Pi3}. 
The RPi also utilizes a 400MHz Broadcom VideoCore IV GPU and has Wi-Fi, Bluetooth, and Ethernet capabilities. 
Similarly, the BB was selected because existing surveys place it between the second and third most popular SBC on the market \cite{2014,2016}.
The BB contains a 1GHz AM3358 ARM Cortex-A8 OSD3358-512M-BAS processor and 512MB of DDR3L SDRAM \cite{BB}. 
The BB also has both Wi-Fi and Bluetooth capabilities. 

When comparing the hardware specifications of both devices, it is important to note two key differences. 
First, while the RPi has nearly twice the SDRAM of the BB, it uses DDR2 SDRAM, which has roughly half the clock speed and transfer rate at 400 to 1,066 MHz and 3,200 to 8,533 MB/s, respectively. 
Additionally, DDR2 SDRAM requires 1.8V to operate, which is relatively high based on modern standards. 
In contrast, the BB, which utilizes  DDR3 `Low Voltage', only requires 1.35V.
A second major difference between the two boards concerns their processor caches. 
For the RPi, the L1 cache level contains 32kB of storage while the L2 cache level contains 512kB of storage. 
The BB has 64K of L1 cache storage that is subdivided into 32K of i-cache and d-cache. 
Additionally, the BB also has 256K of L2 cache storage.
Table \ref{pi_vs_bb} presents the hardware characteristics of these two boards.

%
%
%
\begin{table}
\caption{Specifications of the IoT boards used}
\label{pi_vs_bb}
\begin{tabular}{|c||c|c|}
\hline
\begin{tabular}[c]{@{}l@{}}Board
\end{tabular} & Raspberry Pi 3 (RPi) & BeagleBone Black (BB) \\ \hline\hline
Processor & \begin{tabular}[c]{@{}l@{}}1.2 GHz 64-bit quad-core\\ Broadcom BCM2837 \\ARMv8 \cite{Pi3} 
\end{tabular} & 
\begin{tabular}[c]{@{}l@{}}1GHz TI Sitara AM3359\\ ARM Cortex A8 \cite{BB}\end{tabular} \\ \hline
Instruction Set & ARMv8 & ARMv7 \\ \hline
L1 cache & 32kB & 64K \\ \hline
L2 cache & 512kB & 256 kB \\ \hline
RAM & 1 GB LPDDR2 & \begin{tabular}[c]{@{}l@{}}512MB DDR3L\\ \end{tabular} \\ \hline
Storage & SD & 4GB eMMC, SD \\ \hline
\end{tabular}
\end{table}

According to the IoT Developer Survey conducted by Eclipse in 2018 \cite{OS}, Linux (71.8\%) remains the leading operating system across IoT devices, gateways, and cloud backends. 
As a result, we used Ubuntu Mate on the RPi and the Debian Jessie on BB.
Both operating systems are 32-bit.

In many industrial applications, IoT devices are often under strict energy constraints. 
Under these circumstances, the devices are set to use only absolutely essential services, protocols, and hardware in order to reduce the power consumption of the system \cite{amirtharaj2018profiling}. 
There are many benefits to this system layout including an increase in energy efficiency, a reduction of operating costs for line-powered systems, and an increase in the operating life for battery-powered systems \cite{amirtharaj2018profiling,IoTCostReduction}. 
In order for our energy consumption analyses and models to be realistic, we needed to eliminate the effect of all unwanted components on performance.
Consequently, we disabled all the unnecessary modules that may interfere with energy consumption, such as Bluetooth, Wi-Fi, and Ethernet.
In addition, we used a serial connection (UART) to communicate with the boards.
This method consumes a negligible amount of power, as opposed to a traditional Ethernet or HDMI connection.

\subsection{Power Measurements}
Accurate energy measurement requires enabling and disabling an energy measurement tool based on the operation being performed.
For example, during our experiments, it was necessary to enable and disable energy measurement right before and after the training phase, respectively. 
Therefore, we required a tool that could be directly controlled by the ML program running on the RPi or BB.
To this end, we use the EMPIOT tool \cite{dezfouli2018empiot}, which enables the devices under test to precisely control the instances of energy measurement.
EMPIOT is capable of supersampling approximately 500,000 readings per second to data points streamed at 1KHz.
The current and voltage resolution of this platform are 100$\mu$A and 4mV, respectively, when the 12-bit resolution mode is configured.
The flexibility of this platform allowed us to integrate it with our testbed.

\subsection{Machine Learning Algorithms}

Our paper focuses on supervised image classification.
Supervised learning uses \textit{labelled data}. 
A labeled example consists of an input and output pair. 
The objective of the supervised algorithm is to produce a model that is able to map a given input to the correct output. 
Types of learning tasks that are considered as supervised learning include classification and regression. 
Popular supervised algorithms include Support Vector Machine (SVM) and linear classifiers \cite{ReinforceLearn}.

In order to grasp the impact of the ML algorithm's effect on energy consumption, it is important to test each algorithm on a wide variety of datasets. 
As a result, we selected three algorithms: \textit{SVM}, \textit{logistic regression}, and \textit{k-Nearest Neighbors} (k-NN). 
In addition to being very popular ML algorithms, each has specific strengths and weaknesses that we study in this paper.

SVM operates by mapping input data to a high-dimensional feature space so that data points can be classified, even when the points are not otherwise linearly separable \cite{stanford_SVM}. 
The data is then transformed in such a way that a hyper-plane can separate them.
The objective of SVM is to maximize the distance (margin) from the separating hyper-plane to the support vectors. 
Logistic regression is used to predict the probability of an event by fitting data to a logistic curve \cite{OrdLogReg}. 
It is intended for predicting a binary dependent variable (e.g., $y=1$ or $y=-1)$. 
k-NN classifies a new sample point based on the majority label of its k-nearest neighbors.

SVM can effectively model non-linear decision boundaries while simultaneously being well insulated against pitfalls such as overfitting. 
However, because SVM may utilize multi-dimensional kernels, it is often memory intensive, thereby leading us to believe it would consume large amounts of energy for datasets with greater than two classes. 
Additionally, because SVM was originally designed to be a binary classifier, we wanted to measure the effectiveness of current SVM implementations when applied to datasets with multiple classes \cite{stanford_SVM}. 
Scikit-Learn, the ML library used for our experiments, implements SVM using the `one-vs-one' method to generate its classifiers. 
Using this approach, given $k$ binary classifiers, all pairwise classifiers are evaluated resulting in $k(k-1)/2$ distinct binary classifiers. 
These classifiers, in turn, vote on the test values, which are eventually labeled as the class with the greatest number of votes \cite{svmComplexity}.
 
Similar to SVM, logistic regression is also designed as a binary classifier, though it does not have the same access to non-linear kernels that SVM does. 
While logistic regression generally performs well for datasets consisting of two classes, its performance drops considerably as the number of classes increases. 
For Scikit-Learn's implementation of logistic regression, when the dataset contains a number of classes greater than two, it uses the `one-vs-all' method. 
This involves training a separate binary classifier for each class. As the number of classes increases, so does the processing time per class. 

The k-NN algorithm is among the simplest and most powerful ML algorithms used for classification and regression. 
When the task is classification, k-NN classifies an object by assigning it to the most common class among its k-nearest neighbors. 
While k-NN is generally recognized as a high-accuracy algorithm, the quality of predictions greatly depends on the method used for proximity measurements. 
Consequently, it was important to select an implementation that used an appropriate distance measurement method, especially when the data points occupy multiple dimensions.

\subsection{SciKit-Learn Framework}

For the purposes of our experiments, we used Scikit-Learn, a Python library that includes ML algorithms such as SVM, logistic regression, and k-NN \cite{sklearn}. 
Although Scikit-learn offers options to utilize multi-core processing, only two of our three algorithms implemented in Scikit-Learn can make use of multiple cores, namely k-NN and logistic regression. In order to measure the benefits of multi-core utilization, we recorded data for the RPi and compared it with the data gathered throughout the single-core experimentation. The BB was excluded from this iteration of experimentation because it only has a single core.

Scikit-Learn also includes modules for hyper-parameter tuning and cross validation. One such module is GridSearchCV which performs an exhaustive search in the hyper-parameter space. When “fitting” the model on a dataset, all the possible combinations of parameter values are evaluated and the best combination is retained. This can be computationally expensive, especially if you are searching over a large hyper-parameter space and dealing with multiple hyper-parameters. A solution is to use RandomizedSearchCV, in which a fixed number of hyper-parameter settings are sampled.

\subsection{Training Datasets}
\begin{figure}
\centering
\includegraphics[scale=0.09]{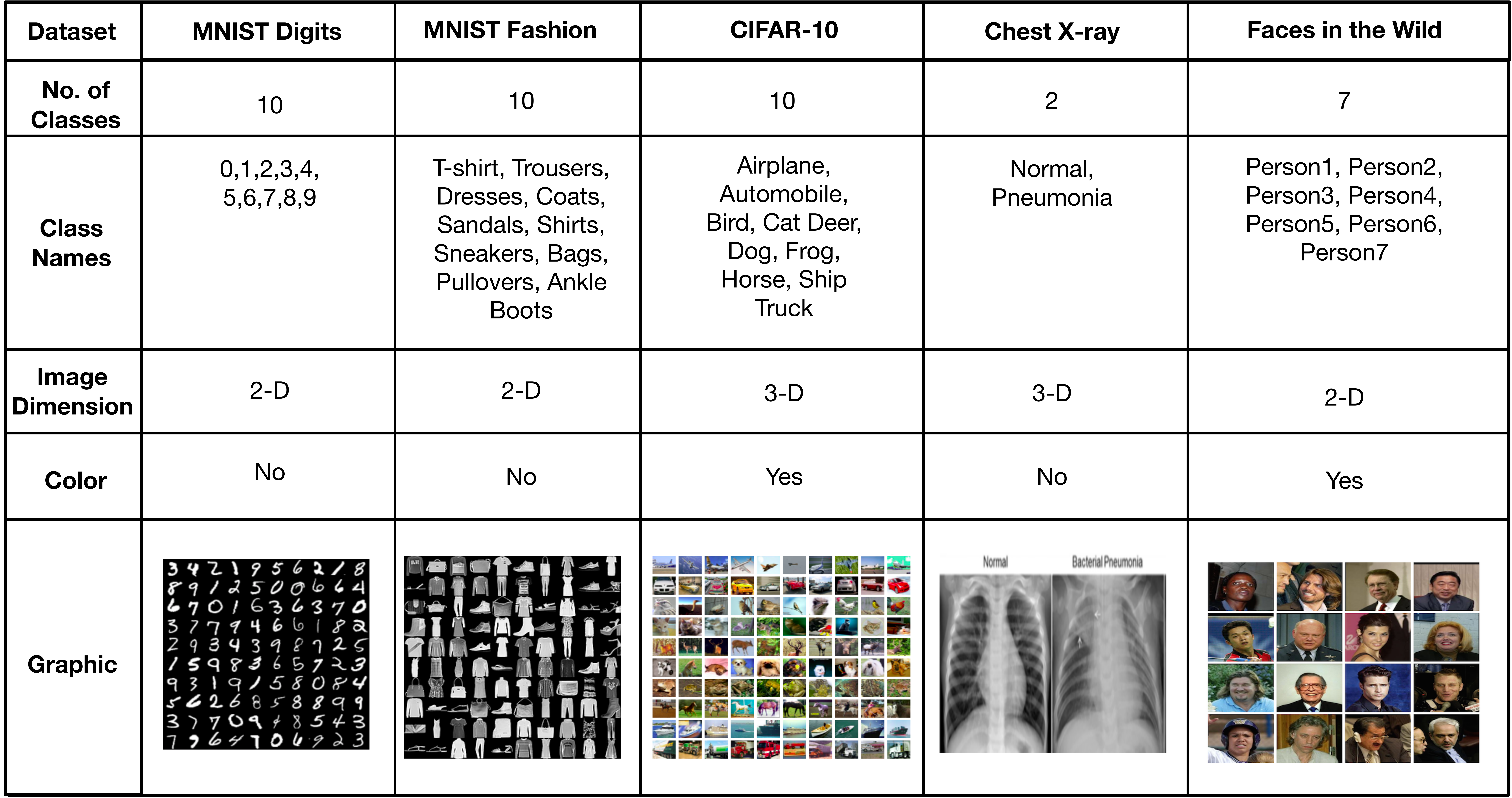}
\caption{Summary of the datasets used in this paper. These datasets enable us to study the impact of various parameters on processing time and energy.}
\label{figure:summ_datasets}
\end{figure}

In order to utilize a diverse range of data, we chose a total of 5 datasets that originally varied in many factors such as image resolution, number of classes, and dataset size. 
We standardized all of these factors in order to fairly compare energy consumption results across multiple datasets. 
No classes overlapped between the datasets, ensuring that our results are pooled from a wide range of test sources. 
The datasets are summarized in the following section and in Figure \ref{figure:summ_datasets}.

\subsubsection{MNIST Digits}

The Modified National Institute of Standards and Technology (MNIST) Digits dataset consists of 70,000 black and white images of handwritten digits \cite{Digits}. 
Each of the digits have been centered in a standardized 28$\times$28 image. 
Each digit corresponds to a separate class resulting in a total of 10 classes. 
This dataset was selected because it is a standard benchmarking dataset.

\subsubsection{Fashion-MNIST}
The Fashion-MNIST dataset was created by researchers at an e-commerce company called Zalando \cite{Fashion}. 
According to the creators, it is intended to serve as a direct drop-in replacement for the original MNIST dataset for benchmarking ML algorithms.  
Similar to the Digits dataset, the Fashion-MNIST dataset consists of 70,000 black and white 28$\times$28 images separated into 10 classes. 
We selected this dataset because it is a more challenging version of the Digits dataset. 
Though widely popular, the Digits dataset has become too elementary with many ML algorithms easily achieving 97\% accuracy. 
Furthermore, most pairs of digits can be distinguished with a single pixel \cite{Fashion}.

\subsubsection{CIFAR-10}

The CIFAR-10 (CIFAR) dataset was created by the Canadian Institute For Advanced Research and consists of 60,000 color images \cite{CIFAR}. 
Each image is 32$\times$32 pixels, and there are a total of 10 classes. 
The classes in this dataset are very diverse ranging from dogs and cats to airplanes and trucks. 
This dataset was selected because it is considered more challenging relative to the other datasets.
The 3-D color images ensure the matrices representing this dataset's images are dense, thus requiring more computation. 
Additionally, because this dataset has 10 significantly different classes and the maximum dataset size is a mere 1500 images, it was intended to represent a scenario where accuracy is low.

\subsubsection{Chest X-Ray}

The Chest X-Ray (CHEST) dataset is provided by Kermany et al. \cite{Xray}.
The dataset contains 5,863 high-resolution greyscale X-ray images divided into two classes: normal and pneumonia.
The images are not square and resolutions are non-uniform. 
This dataset was selected because it only has two classes, which is ideal for SVM and logistic regression.

\subsubsection{Faces in the Wild}

The Labeled Faces in the Wild dataset was created by researchers at the University of Massachusetts Amherst and consists of 13,000 non-square, color images \cite{FITW}.
The images were collected from the web using the Viola-Jones face detector. 
Each of the images is labeled with the name of the person in the picture.
A total of 1,680 individuals pictured within the dataset contain at least two distinct photos. 
This dataset was selected because much like the CIFAR dataset, this dataset contains many classes and color images. 
However, unlike CIFAR, the images within the Faces in the Wild dataset are two dimensional. 



 %
\subsection{Dataset Standardization}
We performed dataset standardization in order to fairly determine the nature of the relationship between certain parameters and energy consumption when executing image classification algorithms.
We began by first selecting 1500 images from each dataset. 
Then, we created four more subsets by reducing the size by 300 images at each iteration.
This yielded subsets of 1200, 900, 600, and 300 images. 
Next, we scaled each of the images from those five subsets into three resolutions: 28$\times$28, 22$\times$22, and 17$\times$17. For 3-D data, the dimensionality of the images were maintained.

\begin{figure}
\includegraphics[scale=0.18]{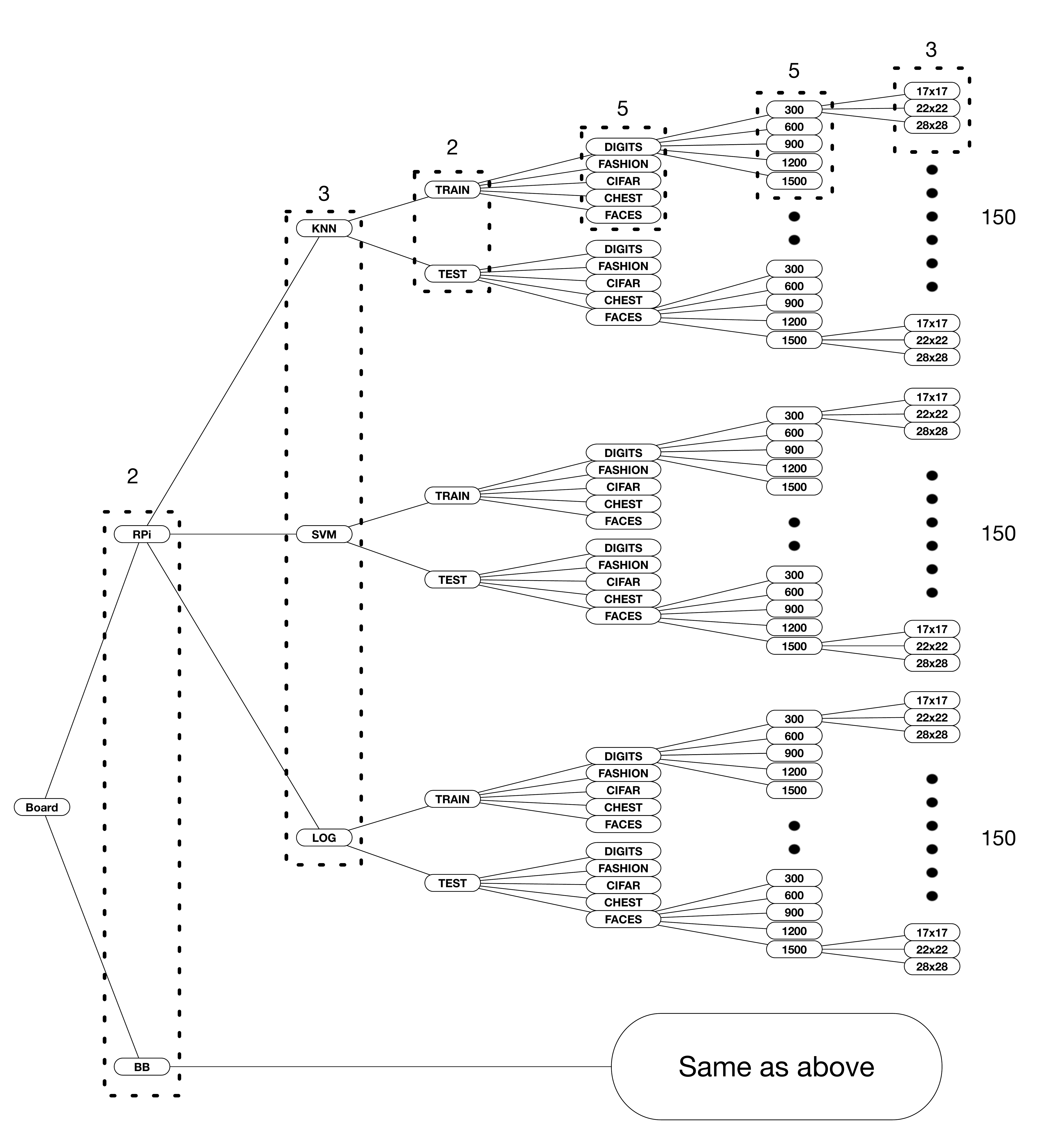}
\caption{Visualization of all the experimental combinations conducted per board. Each experiment is conducted 5 times for a total of 2,250 experiments.}
\label{ExperiementVisual}
\end{figure}

For each iteration of the experiment, we tested a unique combination selected from 3 ML algorithms, 5 datasets, 2 phases, 5 sizes, and 3 resolutions.
This resulted in 450 tests per single complete experiment iteration. 
Furthermore, in order to ensure a reliable measurement, the experiment was executed 5 times for a total of 2,250 experiments. 
Figure \ref{ExperiementVisual} depicts a visualization of the total number of single-core experiments conducted. 

\section{Results and Analysis}
\label{ResultsAnalysisSection}
This section presents and analyzes the different relationships we observed throughout the experiments. 
Specifically, we explore how the algorithm used, as well as various image characteristics, affect energy consumption, processing duration, and accuracy. 

\subsection{Image Resolution}

\begin{figure*}[t!]
    \centering
    \includegraphics[scale=0.21]{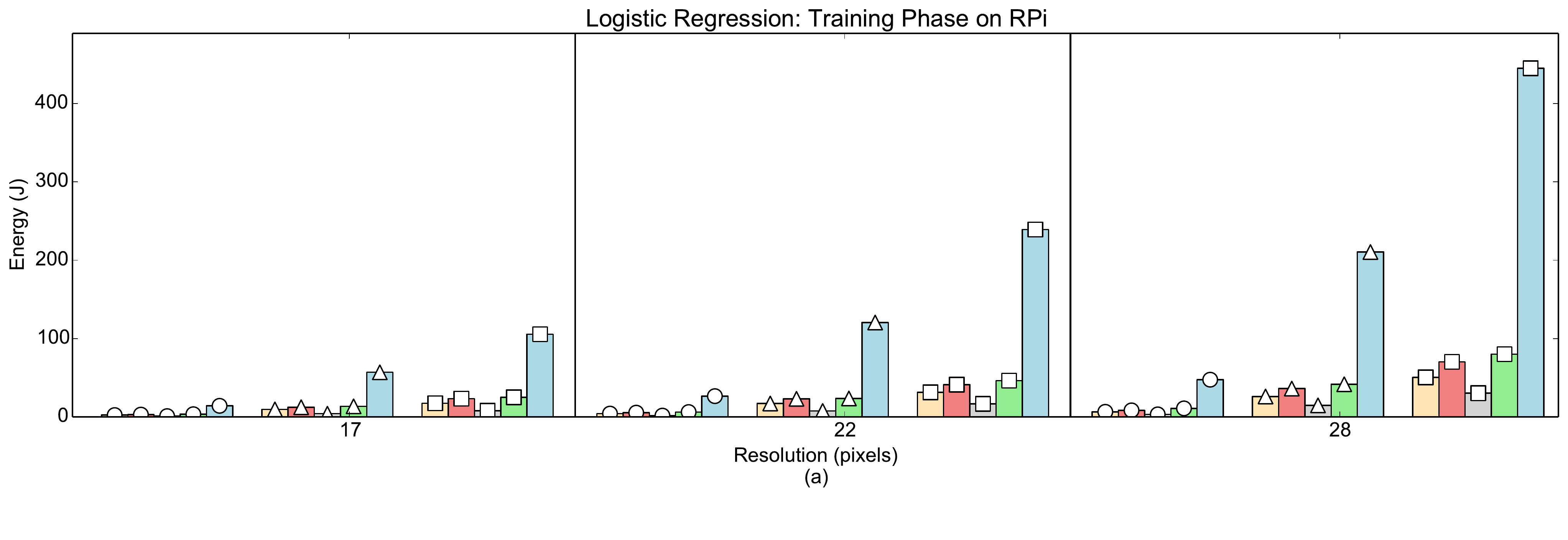}
    \includegraphics[scale=0.21]{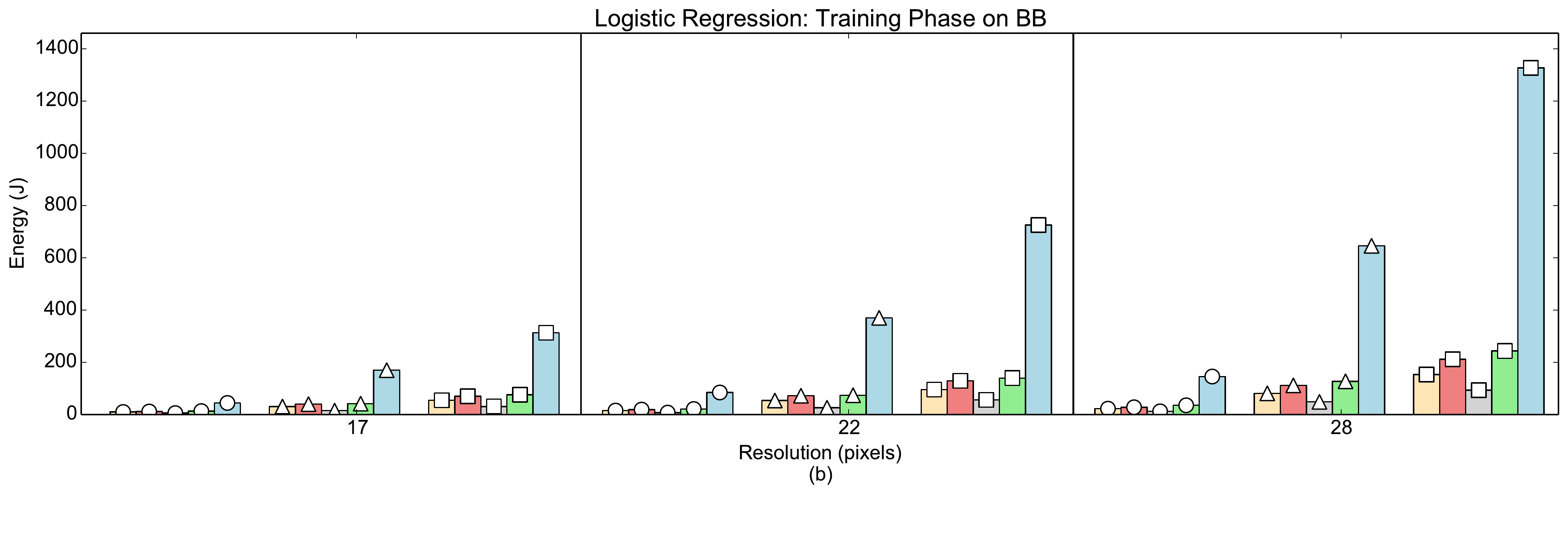}
    \includegraphics[scale=0.25]{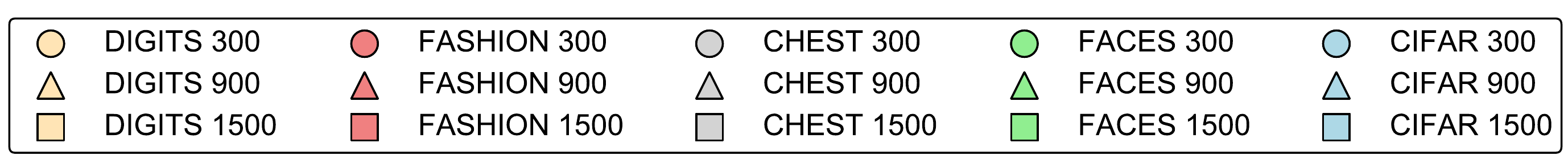}
    \caption{Energy consumption versus resolution during the training phase of Logistic Regression on (a) RPi and (b) BB. Each resolution is separated into three groups representing dataset sizes 300, 900, and 1500. The training phase of logistic regression consumes up to approximately 450J and 1,400J for the RPi and BB, respectively. The trend holds across dataset sizes and boards as demonstrated by the figure.}
    \label{figure:log_train}
\end{figure*}

In this section, we study the effect of image resolution on energy consumption and processing duration. 
Figure \ref{figure:log_train} displays a subset of the collected results for both the RPi and BB during the training phase of logistic regression when the dataset size is held constant.
We observe a linear trend between image resolution and energy consumption for each algorithm during both phases.

\begin{table}
\centering
\caption{The percent increase of energy consumed by CIFAR versus the other datasets when increasing the resolution from 17$\times$17 to 28$\times$28.}
\begin{tabular}{|c||c||c||c|}
\hline
Dataset & Device & Dataset Size & \% Increase \\ \hline\hline
CIFAR   & RPi    & 300          & 550\%            \\ \hline
CIFAR   & RPi    & 900          & 612\%            \\ \hline
CIFAR   & RPi    & 1500         & 636\%            \\ \hline
CIFAR   & BB     & 300          & 446\%            \\ \hline
CIFAR   & BB     & 900          & 583\%            \\ \hline
CIFAR   & BB     & 1500         & 633\%            \\ \hline
\end{tabular}
\label{table:log_train}
\end{table}

A higher resolution implies the device must analyze more pixels in order to classify the image.
An increase in the number of features (pixels) increases the memory consumption and prediction latency.
For a matrix of $M$ instances with $N$ features, the space complexity is in $O(N \times M)$ \cite{sklearn}. 
From a computing perspective, this also means that the number of basic operations (e.g., multiplications for vector-matrix products) increases as well.
Overall, the prediction time increases linearly with the number of features. 
Depending on the global memory footprint and the underlying estimator used, the prediction time may increase non-linearly \cite{sklearn}.
Increasing the memory and the prediction latency directly increases the energy consumption.

\subsection{Dataset Size}
\label{result:dataset}
\begin{figure*}[]
\begin{center}
    \includegraphics[scale=0.35]{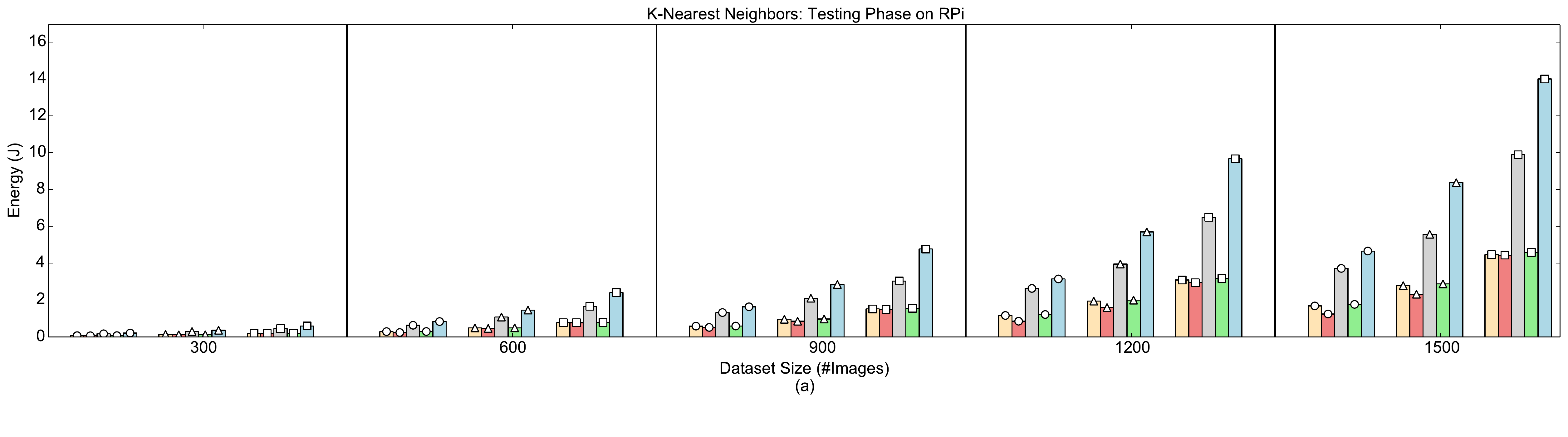}
    \includegraphics[scale=0.35]{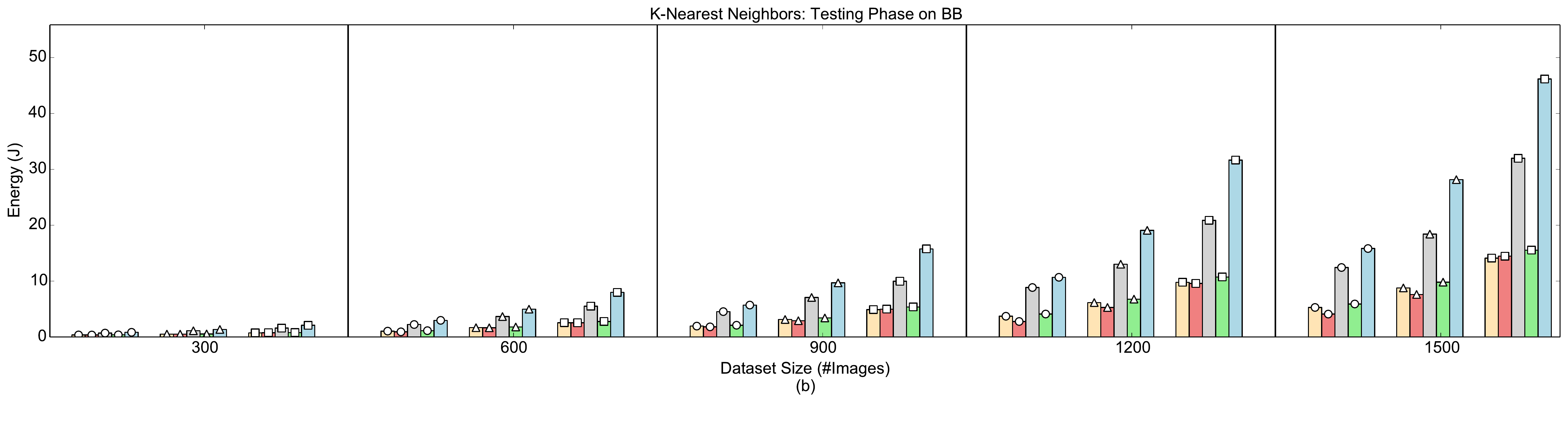}
    \includegraphics[scale=0.3]{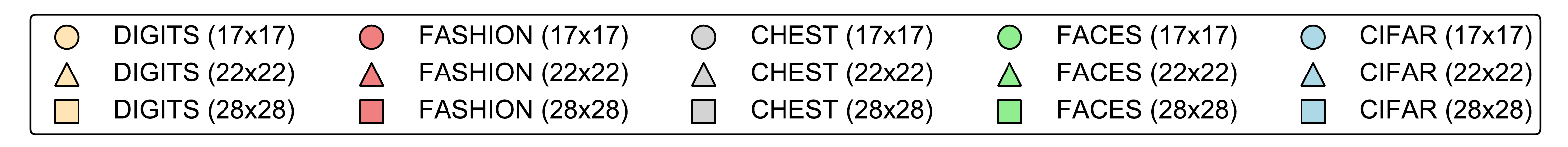}
    \caption{Energy consumption versus dataset size during testing phase of k-Nearest Neighbors on (a) RPi and (b) BB. Each dataset size is separated into three groups representing image resolutions 17$\times$17, 22$\times$22, and 28$\times$28 pixels. Our analysis confirms the quadratic increase of energy consumption versus dataset size.}
    \label{figure:KNN_test}
\end{center}
\end{figure*}

For this experiment, we held all other parameters constant and varied the number of images in the training and testing sets.
Through an analysis of polynomials of varying degrees, we find a quadratic relationship exists between energy consumption and dataset size for each algorithm when the image resolution was held constant.
Figure \ref{figure:KNN_test} is another subset of the collected results that shows this relationship on both hardware platforms during the testing phase of k-NN. Please note the scale difference between Figure \ref{figure:KNN_test} (a) and (b), which highlights the significant difference between the RPi and BB.

\begin{table}
\centering
\caption{The percent increase of energy consumed by CIFAR/CHEST (left column) against the average energy consumed by the other datasets.}
\begin{tabular}{|c||c||c||c|}
\hline
Dataset & Device & Resolution & \% Increase \\ \hline \hline
CIFAR   & RPi    & 17$\times$17         & 104\%            \\ \hline
CHEST   & RPi    & 17$\times$17         & 142\%            \\ \hline
CIFAR   & RPi    & 22$\times$22         & 119\%            \\ \hline
CHEST   & RPi    & 22$\times$22         & 127\%            \\ \hline
CIFAR   & RPi    & 28$\times$28         & 122\%            \\ \hline
CHEST   & RPi    & 28$\times$28         & 111\%            \\ \hline
CIFAR   & BB     & 17$\times$17         & 100\%            \\ \hline
CHEST   & BB     & 17$\times$17         & 128\%            \\ \hline
CIFAR   & BB     & 22$\times$22         & 115\%            \\ \hline
CHEST   & BB     & 22$\times$22         & 114\%            \\ \hline
CIFAR   & BB     & 28$\times$28         & 122\%            \\ \hline
CHEST   & BB     & 28$\times$28         & 106\%            \\ \hline
\end{tabular}
\label{table:KNN_test}
\end{table}
This trend was expected because as the number of images the device has to process during the training phase and classify during the testing phase increases, the longer the device will be running and consuming energy.

\subsection{Image Dimensions}
As expected, datasets with 3-D data (e.g., 28$\times$28$\times$3) generally show higher energy consumption than datasets with 2-D data (e.g., 28$\times$28). 
Both the CIFAR and CHEST datasets had 3-D data and their energy consumption was consistently higher than the remaining datasets. 
In addition, the CIFAR dataset consistently represents the highest energy level because not only does it contain 3-D data, but the matrices representing the images are not sparse.
In order to quantify this increase, we took an average of the energy consumption for all non-CIFAR data and compared it with the energy consumption of the CIFAR data.
The values in Table \ref{table:log_train} are calculated in this way. 
However, in Table \ref{table:KNN_test} we calculated the average energy consumption of the Fashion, Digits, and Faces datasets and compared it with the average energy consumption of CIFAR and CHEST separately.
On average, we found that training a logistic regression model using CIFAR images for dataset sizes of 300, 900, and 1500 consumes 550\%, 612\%, and 636\% more energy, respectively, on the RPi.
We observe a similar trend on the BB, which, under the same circumstances, consumes 446\%, 583\%, and 633\% more energy, respectively.
This is because the CIFAR dataset images contain various colors throughout the entire image as shown in Figure \ref{figure:summ_datasets}, whereas the CHEST images are greyscale and the variance in color is concentrated in the center of the images (the CHEST images generally show a white chest in the center and a black background), thus resulting in sparser matrices.
Scipy, the Python module which Scikit-Learn is built on top of, provides sparse matrix data structures which are optimized for storing sparse data. 
The main benefit of sparse formats is that the space and time complexity decrease significantly.
Specifically, the space complexity decreases because the format does not store zeros. 
Storing a non-zero value requires on average one 32-bit integer position, a 64-bit floating point value, and an additional 32 bits per row or column in the matrix \cite{sklearn}.
Therefore, prediction latency can be dramatically sped up by using a sparse input format since only the non-zero valued features impact the dot product and thus the model predictions.
For example, suppose there are 500 non-zero values in a $10^{3}$ dimensional space. Using Scipy's sparse formats reduces the number of multiply and add operations from $10^{3}$ to 500.

\subsection{Algorithm}
\label{result:algorithm}
Though the image characteristics in isolation affect energy consumption, our results show that the ML algorithm used is consistently the greatest predictor of energy consumption and processing duration. 
This is because these algorithms are designed for specific tasks. 
For example, the CHEST dataset, which contains 3-D images, generally consumes the second highest amount of energy when using SVM and k-NN. 
However, when logistic regression, which is designed for binary classification, is run on the CHEST dataset, we see a dramatic decrease in energy, regardless of its high dimension because the CHEST dataset only has two classes.

In general, we found that logistic regression's training phase consumes significantly more energy than the training phases of the other two algorithms. 
Figure \ref{figure:log_train} (a) and (b) show that the training phase of logistic regression consumes up to approximately 450J and 1,400J for the RPi and BB, respectively.
In comparison, the training phases of k-NN and SVM consume 2J and 100J on the RPi and 9J and 450J on the BB, respectively.
This large discrepancy in energy cost is observed because training a logistic regression model for more than 2 classes involves creating a separate classifier for each class. 
On the other hand, the testing phase for logistic regression consumes significantly less energy than SVM and k-NN because predicting a single image is simply a matter of taking the maximum output across each classifier generated during the training phase. 
This trade-off is an important consideration when determining which algorithm to use for a resource-constrained edge device. 

For k-NN, we also observe quadratic trends, as shown in Figure \ref{figure:KNN_test} for both the RPi and the BB.
During its training phase, k-NN simply stores all the images in the training set. 
Suppose there are $n$ training examples each of dimension $d$. Then the complexity to compute the distance to one example is $O(d)$. 
To find a single nearest neighbor, the complexity is $O(n\times d)$. 
Thus, to find the $k$ nearest neighbors, the complexity is $O(k \times n \times d)$ \cite{KNNComplexity}. As the dataset size increases, the overall complexity increases, which in turn increases the energy consumed.
Therefore, the overall complexity is dependent on the dataset size and the value of $k$.  
Choosing $k$ optimally is not a trivial task. In theory, with an infinite number of samples, as the value of $k$ increases, the error rate approaches the optimal Bayes error rate. 
The caveat being that all $k$ neighbors have to be close to the example. 
However, this is impossible since the number of samples is finite. 
A large $k$ leads to over-smoothed decision boundaries, and a small $k$ leads to noisy decision boundaries. 
For our experiments, we used cross-validation to tune $k$ to a value of 5.

\begin{figure}[t]
\centering
 \includegraphics[width=0.65\linewidth]{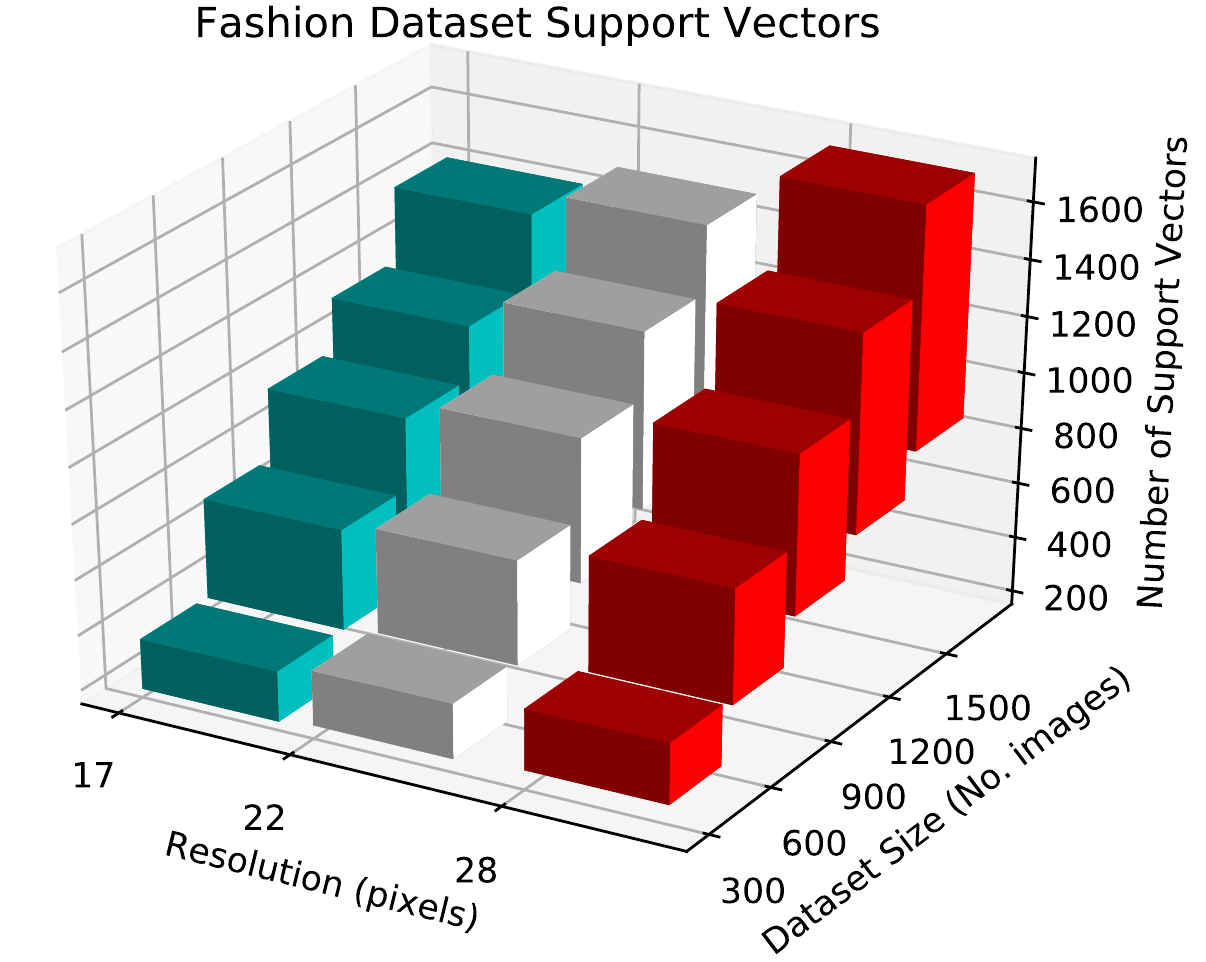}
 \caption{The number of Support Vectors for the FASHION Dataset. Increasing the number of images in a dataset has a greater effect on support vector complexity when compared to increasing image resolution.}
 \label{figure:3D}
\end{figure}

\begin{figure*}[]
\begin{center}
    \includegraphics[scale=0.35]{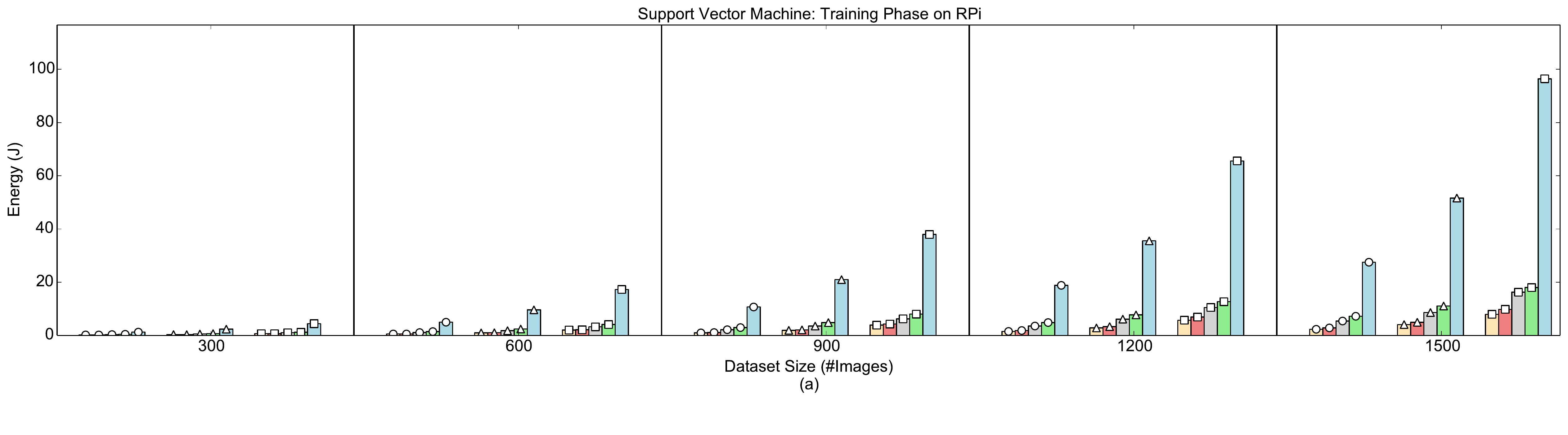}
    \includegraphics[scale=0.35]{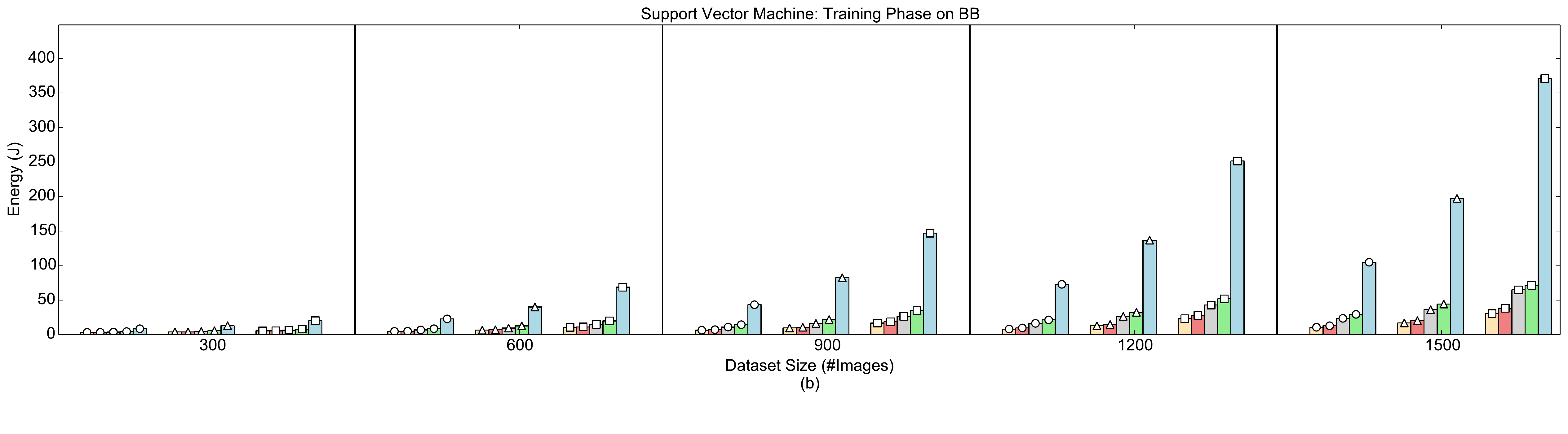}
    \includegraphics[scale=0.3]{legend_rest.pdf}
    \caption{Energy consumption versus dataset size during testing phase of k-Nearest Neighbors on (a) RPi and (b) BB. Each dataset size is separated into three groups representing image resolutions 17$\times$17, 22$\times$22, and 28$\times$28 pixels. The CIFAR and Faces (dense matrix representations) datasets consistently consume more energy during the training phase of the algorithm across all resolutions.}
    \label{figure:SVM}
\end{center}
\end{figure*}

For SVM, the energy consumption depends on the number of support vectors. 
A higher number of support vectors indicates a higher model complexity. 
Our results show that, the processing duration asymptotically grows linearly with the number of support vectors. 
The number of support vectors increases when we increase resolution or dataset size, as demonstrated in Figure \ref{figure:3D}. 
In addition, the non-linear kernel used (radial basis function in Scikit-learn) also influences the latency as it is used to compute the projection of the input vector once per support vector. 
Furthermore, since the core of a SVM is a quadratic programming problem which separates support vectors from the rest of the training data, Scikit-learn's implementation of the quadratic solver for SVM scales between $O(n_{f} \times n_{s}^2) $ and $O(n_{f} \times n_{s}^3)$, where $n_{f}$ is the number of features and $n_{s}$ is the number of samples. 
If the input data is very sparse, $n_{f}$ should be replaced by the average number of non-zero features in a sample vector. 
Figure \ref{figure:SVM} shows that the CIFAR and Faces datasets (which have dense matrices) consistently consume more energy relative to the other datasets during the training phase of the algorithm across all resolutions. 
Figure \ref{figure:SVM} (a) highlights this trend with the CIFAR dataset consuming 636\% more energy than the average consumption of the other four datasets for the resolution of 28.
Table \ref{table:SVM} summarizes the effect of image resolution on energy consumption when using CIFAR dataset.



\begin{table}[]
\centering
\caption{The percent increase of energy consumed between dataset and resolution pairs versus lowest to highest dataset size in Figure \ref{figure:SVM}.}
\begin{tabular}{|c||c||c||c|}
\hline
Dataset & Device & Resolution & \% Increase \\ \hline\hline
CIFAR   & RPi    & 17$\times$17         & 405\%            \\ \hline
CIFAR   & RPi    & 22$\times$22         & 486\%            \\ \hline
CIFAR   & RPi    & 28$\times$28         & 504\%            \\ \hline
CIFAR   & BB     & 17$\times$17         & 290\%            \\ \hline
CIFAR   & BB     & 22$\times$22         & 377\%            \\ \hline
CIFAR   & BB     & 28$\times$28         & 425\%            \\ \hline
\end{tabular}
\label{table:SVM}
\end{table}

\subsection{Time and Accuracy}
In addition to measuring energy consumption, we also measured processing time and classification accuracy. 
Specifically, these studies enable us to offer guidelines for establishing trade-offs between energy consumption and accuracy.
Table \ref{table:deltaRPI} and \ref{table:deltaBB} show the accuracy increase for each algorithm and dataset pair at sizes of 300 and 1500 images. 

\begin{table}[t]
\label{RPI_perc_change}
\centering
\caption{The percent change in time and accuracy when varying between 17$\times$17 and 28$\times$28 image resolutions on the RPi platform. The dataset size is constant.}

\begin{tabular}{|c||c||c||c||c||c|} \hline
Algorithm & Dataset & \begin{tabular}[c]{@{}l@{}}$\Delta t_{300}$\end{tabular} & \begin{tabular}[c]{@{}l@{}}$\Delta a_{300}$\end{tabular} & \begin{tabular}[c]{@{}l@{}}$\Delta t_{1500}$\end{tabular} & \begin{tabular}[c]{@{}l@{}}$\Delta a_{1500}$\end{tabular} \\ \hline\hline

k-NN & DIGITS & 137\% & 0\% & 243\% & -4.80\% \\
k-NN & CHEST & 173\% & -7.4\% & 1.72\% & 1.48\% \\
k-NN & FASHION & 200\% & 0\% & 244\% & 1.85\% \\
k-NN & FACES & 160\% & 0\% & 164\% & 0\% \\
k-NN & CIFAR & 179\% & -12.5\% & 199\% & -23.89\% \\
SVM & DIGITS & 226\% & -4.16\%& 224\% & 2.11\% \\
SVM & CHEST & 200\% & 0\% & 203\% & -0.69\% \\
SVM & FASHION & 226\% & -10\% & 218\% & -4.1\% \\
SVM & FACES & 154\% & 0\% & 136\% &5.31\%\\
SVM & CIFAR & 206\% & -25\% & 233\% & -14\% \\
LOG & DIGITS & 170\% & 0\% & 188\% & -0.71\% \\
LOG & CHEST & 209\% & 7.69\% & 289\% & 0.69\% \\
LOG & FASHION & 184\% & -17.64\% & 202\% & -4.13\% \\
LOG & FACES & 207\% & 0\% & 221\% & 0\% \\
LOG & CIFAR & 236\% & 14.29\% & 324\% & -28.85\% \\ \hline
\end{tabular}
\label{table:deltaRPI}
\end{table}

\begin{table}[t]
\centering
\caption{The percentage increase of processing duration and accuracy when varying image resolution between 17$\times$17 and 28$\times$28 on the BB. The dataset size is constant.}

\scalebox{0.95}{
\begin{tabular}{|c||c||c||c||c||c|} \hline
Algorithm & Dataset & \begin{tabular}[c]{@{}l@{}}$\Delta t_{300}$\end{tabular} & \begin{tabular}[c]{@{}l@{}}$\Delta a_{300}$\end{tabular} & \begin{tabular}[c]{@{}l@{}}$\Delta t_{1500}$\end{tabular} & \begin{tabular}[c]{@{}l@{}}$\Delta a_{1500}$\end{tabular} \\ \hline\hline

k-NN & DIGITS & 18.26\% & 0\% & 120\% & -4.80\% \\
k-NN & CHEST & 40.29\% & -7.4\% & 136\% & 1.48\% \\
k-NN & FASHION & 19.53\% & 0\% & 157\% & 1.85\% \\
k-NN & FACES & 19.44\%& 0\% & 119\% & 0\% \\
k-NN & CIFAR & 48.33\% & -12.5\% & 166\% & -23.89\% \\
SVM & DIGITS & 63.66\% & -4.16\% & 186\% & 2.11\% \\
SVM & CHEST & 75.34\% & 0\% & 178\% & -0.69\% \\
SVM & FASHION & 66.70\% & -10\% & 194\% & -4.09\% \\
SVM & FACES & 77.97\% & 0\% & 143\% & 5.31\% \\
SVM & CIFAR & 132\% & -25\% & 244\% & -14\% \\
LOG & DIGITS & 123\% & 0\% & 178\% & -0.71\% \\
LOG & CHEST & 111\% & 7.69\% & 198\% & 0.69\% \\
LOG & FASHION & 136\% & -17.65\% & 201\% & -4.13\% \\
LOG & FACES & 161\% & 0\% & 218\% & 0\% \\
LOG & CIFAR & 225\% & 14.28\% & 329\% & -28.85\% \\ \hline
\end{tabular}
}
\label{table:deltaBB}
\end{table}

In general, when holding all other factors constant, accuracy does not significantly change when resolution was changed. 
For example, for the RPi, increasing the resolution from 17$\times$17 to 28$\times$28 (by 40\%) while keeping the dataset size constant at 300 images, always resulted in at least double the time and little to no additional increase in accuracy.
Table \ref{table:deltaRPI} shows that the maximum increase in accuracy across all the experiments is approximately 14\% when running logistic regression on a subset of the CIFAR dataset consisting of 300 images.
However, this increases time by 236\%.
We observe that 7 out of the 15 experiments have the same accuracy even when increasing the resolution. 
Additionally, 6 out of the 15 experiments show a decrease in accuracy. 
Thus, 13 out of the 15, or roughly 90\% of the experiments show that there is no additional benefit to using higher-resolution images. 
These accuracy trends, which are identical for the BB, are a critical consideration for many applications. 
As a result, one should generally opt for the reduced resolution.

The increases in energy consumption associated with opting for higher resolution can instead be allocated to increasing the dataset size. 
This could lead to an increase in accuracy.
Figures \ref{figure:KNN_test} and \ref{figure:SVM} demonstrate that choosing the higher resolution is roughly equivalent to choosing the lower resolution at a higher dataset size. 
For example, in Figure \ref{figure:SVM}, selecting a dataset size of 300 images at a resolution of 28x28, consumes approximately the same amount of energy as a dataset of 600 images at a resolution of 17$\times$17.
Again in Figure \ref{figure:SVM}, we observe that selecting a dataset size of 600 images at a resolution of 28 consumes approximately the same amount of energy as that of a dataset of 1200 images at a resolution of 17$\times$17.

\subsection{Multi-core versus Single-core}
\begin{figure}
\begin{center}   
    \includegraphics[width = 0.65\linewidth]{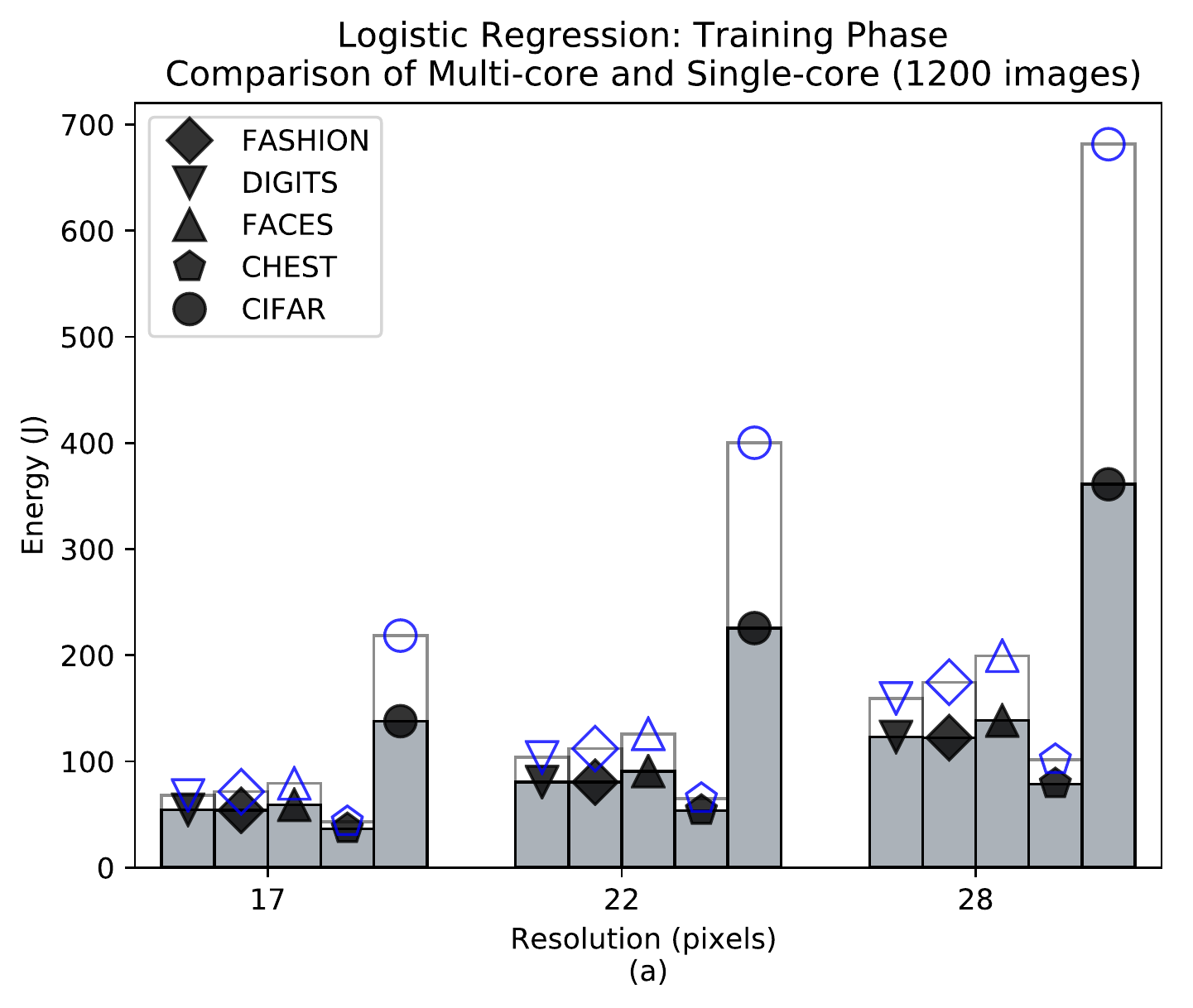} \hfill
\end{center}
\begin{center}
    \vspace{0.5cm}
    \includegraphics[width = 0.65\linewidth]{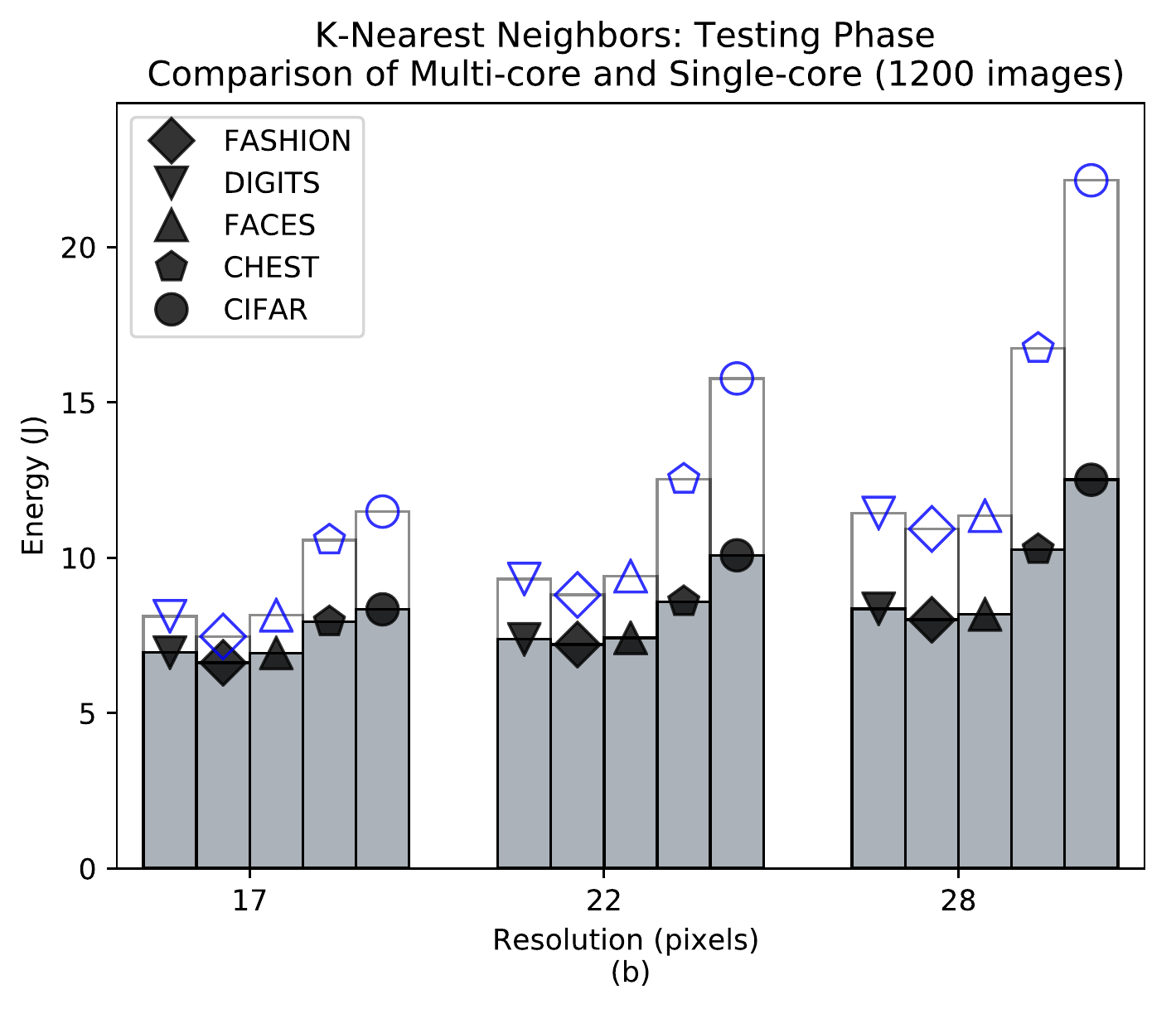} \hfill
    \caption{Comparison of energy consumption during the training phase of logistic regression using multi-core functionality (dark fill) and single-core functionality (no fill).}
    \label{figure:core_data_energy}
    \end{center}
\end{figure}

\begin{figure}
    \centering
    \includegraphics[width = 0.65\linewidth]{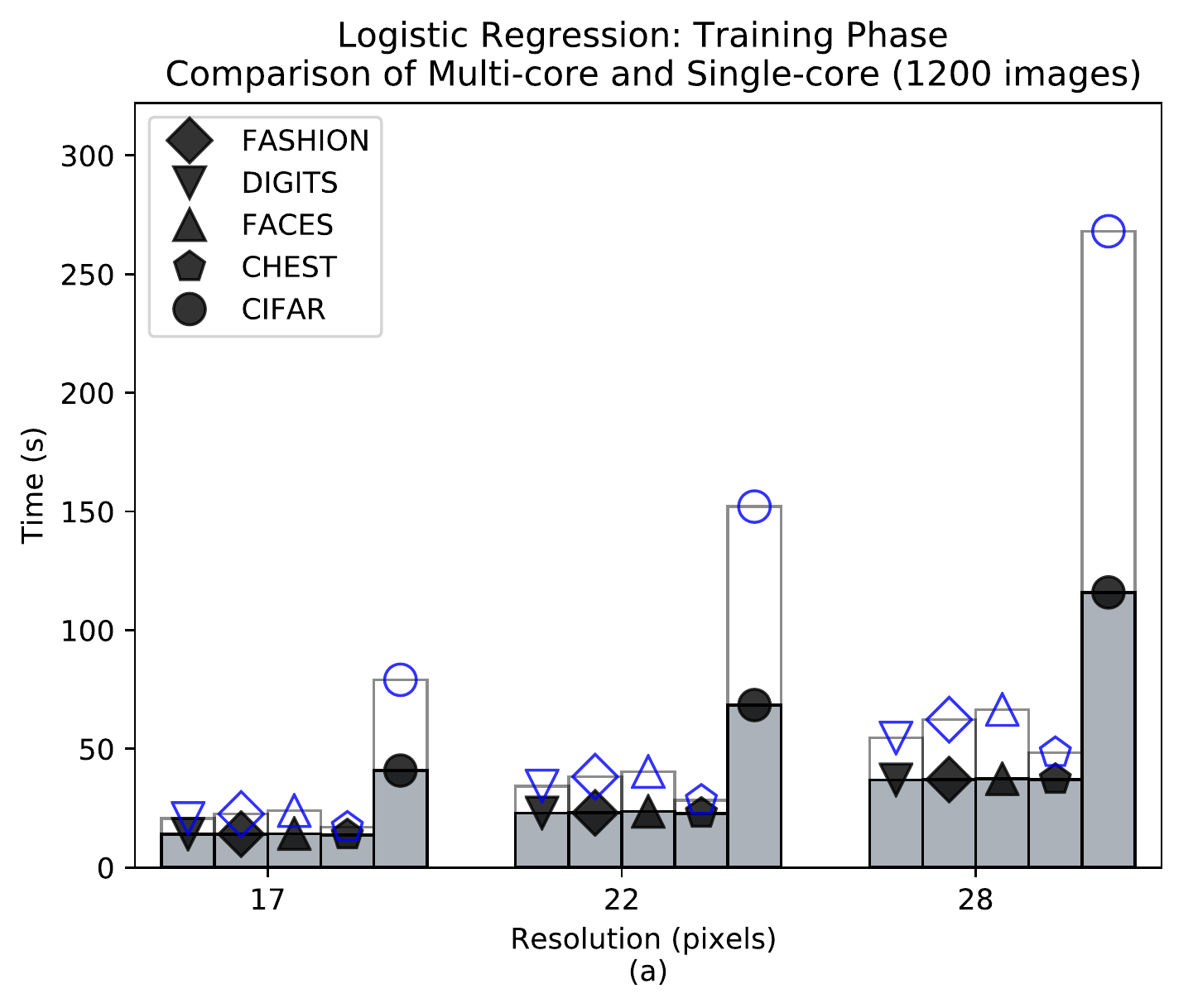}
    \vspace{0.5cm}
    \includegraphics[width = 0.65\linewidth]{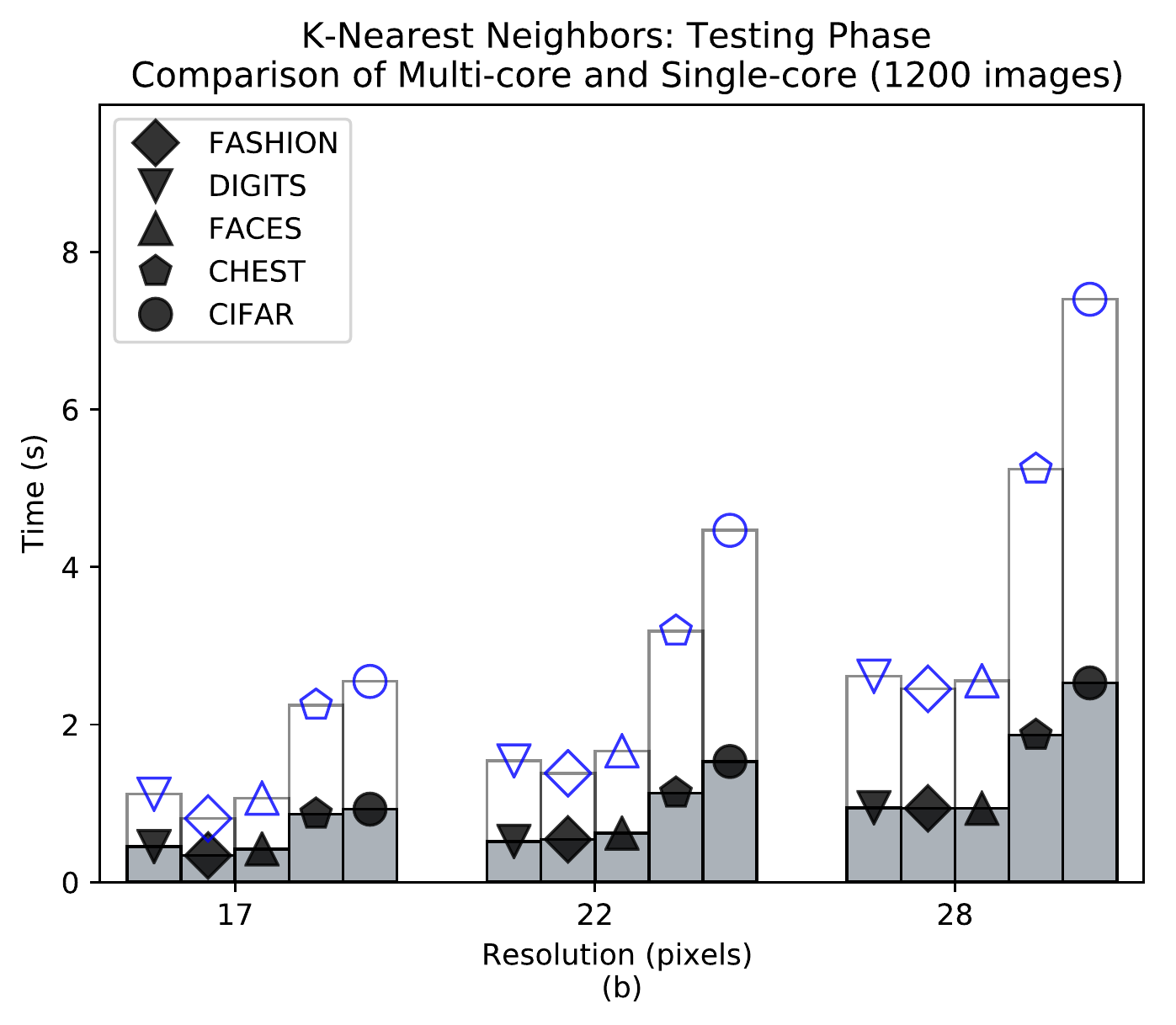}
    \caption{Comparison of processing time during the training phase of logistic regression using multi-core functionality (dark fill) and single-core functionality (no fill).}
    \label{figure:core_data_time}
\end{figure}

To quantitatively determine how the usage of multiple cores affects the energy consumption and processing time, we executed k-NN and logistic regression (the algorithms capable of multi-core processing) using all four cores on the RPi. 
Figures \ref{figure:core_data_energy} and \ref{figure:core_data_time} demonstrate the differences between multi-core and single-core processing time and energy consumption for the testing phase of k-NN and the training phase of logistic regression. 
For both time and energy, there is a significant gap between using multi-core and single-core. 
On average, when utilizing multi-core functionality, the processing time for k-NN and logistic regression was reduced by 70\% and 42\%, respectively. 
Using multiple cores also translated in a 63\% and 60\% decrease in energy consumption for the same two algorithms, respectively.

\subsection{Design Considerations}
In this section, we present our main observations regarding the effect of hardware on performance as well as a set of design guidelines to create real-time IoT systems for the purpose of image classification.

\subsubsection{Hardware}
While the trends we identified in the previous sections are consistent across the two hardware platforms, it is important to note the dramatic differences in their individual energy consumption.
The RPi not only has a CPU that is 20\% faster than the BB, but it also boasts nearly twice as much RAM. 
This variance in hardware results in the RPi completing the experiments much faster than the BB. 
Consequently, because the BB had to run longer to complete each task, we observe that on average it consumes more energy. 
This conclusion is best demonstrated by Figure \ref{figure:log_train} (a) and (b), which display the training phase of logistic regression when a single core is used. 
The RPi, which could generally train a logistic regression model for more than two classes in less than 4 minutes, consumed up to 450J. 
By contrast, the BB, which generally took 15-19 minutes to train a logistic regression model, consumed up to 1,400J. 
This trend is consistent across all algorithms. 
More importantly, the RPi can achieve significantly higher performance compared to the BB, when the ML algorithm utilizes all the available four cores.
In particular, the fast growth of low-cost, multi-core processors justifies their adoption at the IoT edge to lower energy consumption and enhance real-time processing.

\subsubsection{Guidelines}
The following guidelines are primarily concerned with energy-performance trade-offs.
Foremost, we observe that for small datasets it is rarely beneficial to increase image resolution.
In most cases, doing so is detrimental to the accuracy of the model and in all cases there is a significant increase in energy consumption as a result of additional pixel analysis.
Second, we suggest to constrain dataset size to a minimum.
While increasing the training set to tens of thousands of images would likely result in higher accuracy, for the small set increments associated with current IoT systems and traditional ML algorithms, adding additional images is not guaranteed to provide significant benefit.
However, similar to increasing image resolution, increasing the dataset size or dimensionality always translates in higher energy consumption.
Third, we suggest that images be captured in methods suited for sparse matrix representations so that they may benefit from the optimizations associated with sparse matrix formats.
It should be noted that in addition to enhancing performance, these methods also can be applied to improve user privacy.
For example, low-resolution and sparse images that do not reveal the person's identity could be captured by thermal cameras to achieve low-power and real-time activity recognition \cite{Whitehouse}.

\section{Modeling and Predicting Energy Consumption}
\label{ModelingRFSection}

Our experimentation provided us with a sizable amount of data that can be used to model and predict the energy consumption.
In this section, we utilize three statistical analysis techniques, discuss the drawbacks and benefits of each, and compare their performance in terms of prediction accuracy.

\subsection{Random Forest Regression}

In random forest regression, a multi-variate input $x$ is used to estimate a continuous label $y$ using the probability density function $p(y|v)$, where $y \in Y\subseteq \mathbb{R}^{n}$ \cite{RF1}. 
In our case, the input contains the following features: device, resolution, number of images, color, number of dimensions, algorithm, and phase of the algorithm. 
All the features were coded to be categorical. 
For a feature with $n$ possible values, we created $n-1$ binary variables to represent it. 
For example, for 3 possible resolutions, we created 2 columns, $r_{1}$ and $r_{2}$, such that for a resolution of 17$\times$17, $r_{1}=0$ and $r_{2}=0$.  
The full encoding for the features is summarized in Table \ref{EncodingFormat}.
\begin{center}
\begin{table}
\centering
\caption{Random Forest Regression Encoding Format}
\label{EncodingFormat}
\begin{tabular}{ |c||c||c|  }
 \hline
 \multicolumn{3}{|c|}{Feature Encoding} \\
 \hline
 Feature & Possible Values & \#Columns\\
 & & to Represent\\
 \hline\hline
 Resolution   & 17$\times$17, 22$\times$22, 28$\times$28   &2\\ \hline
 \#Images &   300,600,900,1200,1500  & 4  \\\hline
 \#Classes & 2,7,10 & 2\\\hline
 Phase    & Train, Test & 1\\\hline
 Color &   Yes, No  & 1\\\hline
 Algorithm& k-NN, SVM, LOG  & 2   \\\hline
 Device & RPi, BB & 1\\
\hline
\end{tabular}
\end{table}
\end{center}
Constructing a random forest model can be broken down into the following steps:
\begin{itemize}
  \item [--] Using random sampling, choose $N$ samples from the training data. 
  \item [--] For each sample, randomly choose $K$ features. Construct a decision tree using these features. 
  \item [--] Repeat steps 1 and 2 for $m$ times to generate $m$ decision tree models. 
\end{itemize}

The above process results in a random forest of $m$ trees. 
To predict the $y$ output of a new query point, pass the input to each of the $m$ trees. 
For regression, the output is the average of $m$ decision tree outputs. 
For classification, the output is the majority class label of the $m$ decision tree outputs \cite{RF2}.
%
%

\subsection{Gaussian Process and Linear Regression}
\label{sec_stat_anal_meth}
In addition to the random forest model, we also evaluate the accuracy of linear regression and Gaussian Process (GP). 
Linear regression is the most common predictive model to identify the relationship among an independent variable and a dependent variable \cite{LINREG}. The multiple linear regression line is fit to the data such that the sum of the squared errors is minimized.
A Gaussian Process defines a distribution over functions which can be used for regression. The main assumption of GP is that the data is sampled from a multivariate Gaussian distribution. The function-space view of GP shows that a GP is completely specified by its mean function and co-variance function.

\subsection{Results and Discussion}
To assess and understand how the proposed prediction model performs on datasets that are not part of the original data, we chose two new datasets and collected data on their energy consumption. 
The first dataset was drawn from Caltech-256 and was chosen because it contains a more challenging set of object categories \cite{caltech256}.
From this dataset, we drew 10 separate, mutually exclusive classes.
The images within these sub-datasets are in color and of varying image resolutions. 
The second verification dataset contains images of flowers \cite{Flowers}. 
We chose this dataset because it contains five classes, which is a characteristic the random forest was not trained to predict. 
The images within this dataset are also in color and of varying image resolutions. 
Following the same experimental protocols, we separated images from each of the datasets into the standard dataset sizes and resolutions.
For both datasets, we only chose the 3-D images because our prior experiments demonstrated that the datasets with 3-D images had the highest variations in energy consumption.
The characteristics of both datasets are summarized in Table \ref{validationDS}.
\begin{center}
\begin{table}[t!]
\centering
\caption{Characteristics of the Verification Datasets}
\label{validationDS}
\begin{tabular}{ |c||c||c||c|}
 \hline
 Dataset & \#Classes & \#Dimensions & Color\\
 \hline \hline
 Caltech-256 & 10 & 3 & Yes\\
 Flowers & 5 & 3 & Yes\\
 \hline
\end{tabular}
\end{table}
\end{center}

Table \ref{R2COMP} demonstrates that linear regression is least successful at prediction.
This algorithm shows 4.2x and 5.9x lower $R^2$ values compared to GP and random forest, respectively, especially when we attempted to extrapolate beyond the range of the sample data.
This is because linear regression requires assumptions that are invalidated by our data.
Specifically, linear regression, as the name suggests, requires the relationship between the independent and dependent variables to be linear, which is not necessarily guaranteed by our data. 
A GP also requires certain assumptions about the data. For example, a GP requires that all finite dimensional distributions have a multivariate distribution. Specifically, each observation must be normally distributed. 
Considering our application, for a specific observation $\mathbf{x_{i}}$, where $\mathbf{x_{i}} = (x_{1},...,x_{n})$, $x_{1}=0$ if the image is 2-D, and $x_{1}=1$ if it is 3-D. 
This immediately precludes the normality assumption imposed by a Gaussian model. 

We used k-fold cross validation with $k=10$ to select the random forest that produces the maximum R-squared value and minimum RMSE. 
Performing k-fold cross validation with $k=5$ or $k=10$ has been shown empirically to yield test error rate estimates that suffer neither from excessively high bias nor from very high variance \cite{STAT}.
%
%
The random forest predicts the energy consumption of the Caltech dataset with a R-squared value of 0.95 and a R-squared value of 0.79 for the Flowers dataset. 
The coefficient of determination, known as R-squared, can be interpreted as the percent of the variation in $y$ that is explained by the variation in the predictor $x$.
A value of 1 for R-squared indicates all of the data points fall perfectly on the regression line which means the predictor $x$ accounts for all of the variation in $y$.
In general, the closer the R-squared is to 1.0, the better the model.

The random forest model is capable of ameliorating the high error rates of the two previous models because it can capture the non-linearity in the data by dividing the space into smaller sub-spaces depending on the features. 
In addition, there is no prior assumption regarding the underlying distribution of the features. As a result of the bias-variance trade off, increasing the number of trees, as well as their depth, is almost always better. More trees reduce the variance; deeper trees reduce the bias. In practice, there are millions of training examples which will make it unrealistic and infeasible to train a random forest without limiting its depth. In fact, not controlling or tuning these parameters can lead to fully grown and un-pruned trees.

\begin{table}[t!]
\caption{Hyper-parameters and Tuned Values of the Random Forest}
\begin{center}
\begin{tabular}{|l|c|}
\hline
Description & Value \\
 \hline \hline

The number of trees in the forest & 800 \\ \hline
\begin{tabular}[c]{@{}l@{}}The number of features to consider \\ when looking for the best split\end{tabular} & sqrt(\#features) \\ \hline
The maximum depth of the tree & 90 \\ \hline
\begin{tabular}[c]{@{}l@{}}Whether bootstrap samples are \\ used when building trees\end{tabular} & True \\ \hline
\begin{tabular}[c]{@{}l@{}}The function to measure the quality \\of a split\end{tabular} & MSE \\ \hline
 \begin{tabular}[c]{@{}l@{}}The minimum number of samples \\required to be at a leaf node\end{tabular} & 4 \\ \hline
 \begin{tabular}[c]{@{}l@{}}The minimum number of samples \\required to split an internal node\end{tabular} & 10 \\ \hline
\end{tabular}
\end{center}
\label{hyperparams}

\end{table}

Some important hyper-parameters that require tuning are included in Table \ref{hyperparams}.
To tune our hyper-parameters, we created a grid of values for each parameter and conducted a randomized grid search with cross validation to choose the optimal values. 
The most important hyper-parameter is the number of features to test at each split. 
Examining feature importance enables us to identify the impact of each feature on the accuracy of the model.
This information can also be used to develop new features, exclude noisy features, and inform another model such as a neural network.

\begin{figure}
\centering
\includegraphics[width=0.9\linewidth]{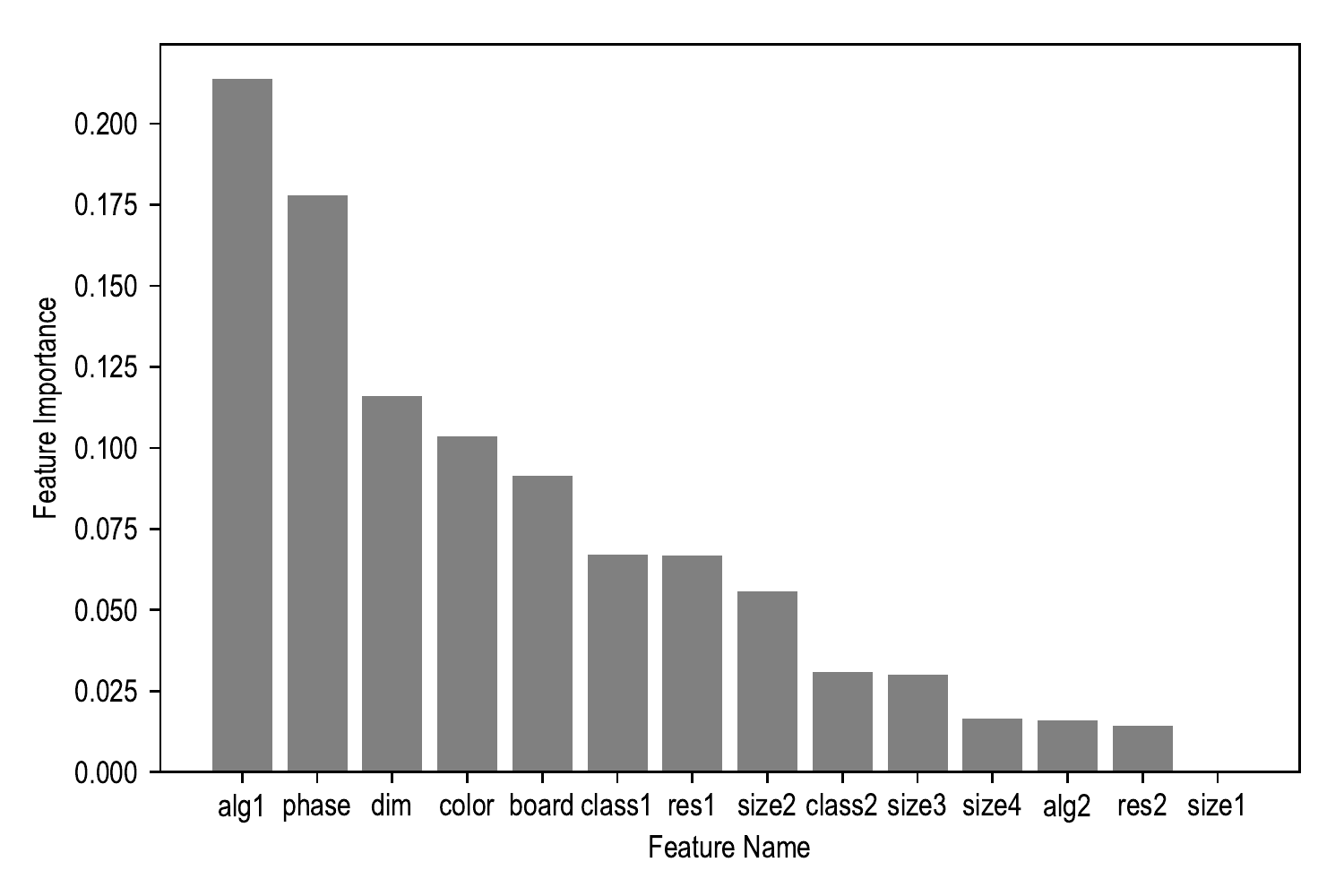}
\caption{Feature importance (a.k.a., `Gini importance' and `mean decrease impurity') for the random forest model. The feature importance is defined as the total decrease in node impurity averaged over all trees of the ensemble.}
\label{figure:feat_importance}

\end{figure}

When determining which feature to split a tree on, the model considers a random permutation of the features.
The number of features to consider was originally suggested as the total number of features divided by 3.
However, recently, it has been empirically justified to use the number of features. 
When using large datasets, it might not be practical to consider thousands of features at each split. 
As a result, it has also been suggested to use the square root of the number of features or the log of the number of features. 
Our model contains 14 features as described in Table \ref{EncodingFormat}. 
Their importance is highlighted in Figure \ref{figure:feat_importance}. 
We found that \textit{size2} (the second column of the encoding for size) is computed to have about 5x higher importance than \textit{size4}, while their `true' importance is very similar. When the dataset has two (or more) correlated features, then from the point of view of the model, any of these correlated features can be used as the predictor, with no concrete preference of one over the others.

When using the impurity-based ranking (for determining which feature to split on), once one of the features is used at the split, the importance of other correlated features is significantly reduced. 
When our goal is to reduce over-fitting, it makes sense to remove features that are mostly duplicated by other features. 
However, this might lead one to the incorrect conclusion that one of the variables is a stronger predictor compared to other features in the same group when in fact, they share a very similar relationship with the response variable. 
It is important to note the difficulty of interpreting the importance of correlated variables is not specific to random forests, but applies to most model-based feature selection methods.



We examined RMSE to quantitatively measure the performance of the random forest model. 
To place it in the context of the data the model is predicting, the RMSE must be normalized. One such method of normalization involves dividing the RMSE by the range.
Table \ref{RMSECOMPARISON} highlights the range, RMSE, and normalized RMSE, separated by algorithm and phase for each dataset. Datasets are coded as follows: original (O), Flowers (F), and Caltech (C).


In order to evaluate the random forest's performance in terms of the two validation datasets, we took an average of their normalized RMSEs for each algorithm and phase.
The average normalized RMSE for k-NN, SVM, and logistic regression during the training and testing phases are as follows: 72.1\%, 11.6\%, 8.4\%, 15.8\%, 18.7\%, and 69.5\%, respectively.
Among these values, SVM has the lowest average error rate at 12.1\%.
This error is approximately 3.4x less than what was exhibited by k-NN and logistic regression, on average. 
The model performs poorly for k-NN's training phase and logistic regression's testing phase on the verification datasets because of the extreme polarity and variation in the data for these algorithm and phase combinations. 
For example, for the Flowers dataset, the maximum value for logistic regression testing is 6.61J, while the minimum is 0.13J. This variation may cause the random forest to poorly predict this configuration. 
As anticipated, the random forest outputs the lowest prediction accuracy for the Flowers dataset because the training set for the random forest did not include a dataset with five classes.

\begin{center}
\begin{table}[t!]
\centering
\caption{$R^{2}$ comparison}
\label{R2COMP}
\begin{tabular}{ |c||c||c||c|  }
 \hline
 ML Model & Flowers & CALTECH-256 & Original Datasets\\
 \hline\hline
 Linear Regression & 0.34 & 0.30 & -0.15\\
 Gaussian Process & 0.50 & 0.63 & 0.49\\
 Random Forest & 0.79 & 0.95 & 0.74\\
 \hline
\end{tabular}
\end{table}
\end{center}

\begin{table}[]
\centering
\caption{RMSE of the predictions achieved by the random forest model when predicting the energy consumption of k-NN, SVM, and LOG.}
\label{RMSECOMPARISON}
\begin{tabular}{|c||c||c||c||c||c|}
\hline
Algorithm & Phase & Dataset & RMSE & Range & N\_RMSE \\ \hline \hline
\rowcolor{Gray}
k-NN & Train & O & 0.162 & 4.409 & 0.036 \\ \hline
k-NN & Train & F & 8.518 & 11.55 & 0.737 \\ \hline
k-NN & Train & C & 8.400 & 11.91 & 0.705 \\ \hline
\rowcolor{Gray}
k-NN & Test & O & 1.389 & 14.14 & 0.098 \\ \hline
k-NN & Test & F & 5.464 & 45.28 & 0.120 \\ \hline
k-NN & Test & C & 4.989 & 44.43 & 0.112 \\ \hline \hline
\rowcolor{Gray}
SVM & Train & O & 27.508 & 196.81 & 0.139 \\ \hline
SVM & Train & F & 30.980 & 348.01 & 0.089 \\ \hline
SVM & Train & C & 26.509 & 334.57 & 0.079 \\ \hline
\rowcolor{Gray}
SVM & Test & O & 1.561 & 12.53 & 0.124 \\ \hline
SVM & Test & F & 5.004 & 30.19 & 0.165 \\ \hline
SVM & Test & C & 4.399 & 28.92 & 0.152 \\ \hline \hline
\rowcolor{Gray}
LOG & Train & O & 170.54 & 1305 & 0.130 \\ \hline
LOG & Train & F & 181.08 & 776.67 & 0.233 \\ \hline
LOG & Train & C & 157.74 & 1117.46 & 0.141 \\ \hline
\rowcolor{Gray}
LOG & Test & O & 0.021 & 0.248 & 0.086 \\ \hline
LOG & Test & F & 4.465 & 6.476 & 0.689 \\ \hline
LOG & Test & C & 4.379 & 6.241 & 0.701 \\ \hline
\end{tabular} 
\end{table}

\section{Conclusion}
\label{ConclusionSection}
As IoT systems become increasingly powerful, edge computing has become more practical compared to offloading data processing to cloud platforms.
This trend has unlocked enormous potential in sectors focused on real-time computing, as it allows IoT systems to quickly and reliably process data while consuming lower energy overall.
This is particularly useful for IoT systems involved in image classification, where the timely processing of data is critical for prompt decision making.
Our experiments sought to explore the relationships between energy consumption, processing duration, and accuracy, versus various parameters including dataset size, image resolution, algorithm, phase, and hardware characteristics. 
We benchmarked two IoT devices in order to reliably identify and study the parameters affecting energy consumption.
Our studies show distinct quadratic and linear relationships between dataset size and image resolution with respect to energy consumption. 
Choosing a lower resolution speeds up the execution time and reduces energy consumption significantly, while maintaining the accuracy of a model trained on a higher resolution. 
If even a slight change in accuracy is crucial for a given system, then selecting a lower resolution frees energy that can be allocated towards increasing dataset size.

In addition to being a lengthy process, energy profiling requires an accurate and programmable power measurement tool. 
Consequently, we propose a random forest model to predict energy consumption given a specific set of features.
While we demonstrated that our model can predict energy consumption with acceptable accuracy for inputs with previously unseen characteristics, it relied on most of the remaining parameters being similar. 
To address this concern, this model would greatly benefit from additional training on datasets with completely different characteristics. 
Additionally, we searched a small hyper-parameter space when tuning the random forest. 
Thus, future work could be focused on testing more values.
In order to further increase the versatility of the model, future work could be focused on collecting data from additional hardware devices to reflect the diversity in the IoT community.
Additionally, in its current state, this model can only be used statically.
Should a user rely on this model for a specific task, prior to deployment, they would need to test each algorithm against the task to find the impact of each on energy consumption.
Future work may also focus on creating a system that dynamically changes the algorithm based on the real-time data returned from the model.
Furthermore, future work could also focus on adding more ML algorithms to the model. 
Pairing this concept with the addition of a model trained on multiple hardware devices would result in a robust system capable of performing tasks optimally and making the best use of its limited computational resources, especially energy.

\section*{Acknowledgment}
This research has been partially supported by the Latimer Energy Lab and Santa Clara Valley Water District (grant\# SCWD02).
This project involves the development of a flood monitoring system where Linux-based wireless systems, which rely on solar or battery power, capture images for analysis using ML to classify and report the debris carried by rivers and streams.

\bibliographystyle{IEEEtran}
\bibliography{reference}

\end{document}